\documentclass{article} 
\usepackage{iclr2019_conference,times}
\usepackage[T1]{fontenc}
\usepackage{amsmath,amsfonts,bm}
\usepackage{graphicx}
\usepackage{tabularx}

\usepackage{amsmath,amsfonts,bm}

\def\eqref#1{equation~\ref{#1}}

\def\1{\bm{1}}

\def\vm{{\bm{m}}}

\def\vx{{\bm{x}}}

\def\mA{{\bm{A}}}

\def\mM{{\bm{M}}}

\def\mU{{\bm{U}}}
\def\mV{{\bm{V}}}
\def\mW{{\bm{W}}}
\def\mX{{\bm{X}}}

\DeclareMathAlphabet{\mathsfit}{\encodingdefault}{\sfdefault}{m}{sl}
\SetMathAlphabet{\mathsfit}{bold}{\encodingdefault}{\sfdefault}{bx}{n}

\def\sR{{\mathbb{R}}}

\DeclareMathOperator*{\argmin}{arg\,min}

\usepackage{hyperref}
\usepackage{url}

\title{Detecting Memorization in ReLU Networks}

\author{Edo Collins, Siavash Arjomand Bigdeli, Sabine S\"usstrunk \\
	School of Computer and Communication Sciences, EPFL\\
	\texttt{\{edo.collins,siavash.bigdeli,sabine.susstrunk\}@epfl.ch} \\
}

\newcolumntype{Y}{>{\centering\arraybackslash}X}

\iclrfinalcopy 
\begin{document}

	\maketitle
	
	\begin{abstract}
		We propose a new notion of `non-linearity' of a network layer with respect to an input batch that is based on its proximity to a linear system, which is reflected in the non-negative rank of the activation matrix.
		We measure this non-linearity by applying non-negative factorization to the activation matrix.
		Considering batches of similar samples, we find that high non-linearity in deep layers is indicative of memorization.
		Furthermore, by applying our approach layer-by-layer, we find that the mechanism for memorization consists of distinct phases.
		We perform experiments on fully-connected and convolutional neural networks trained on several image and audio datasets. Our results demonstrate that as an indicator for memorization, our technique can be used to perform early stopping.
	\end{abstract}

	\section{Introduction}
	
	A fundamental challenge in machine learning is balancing the bias-variance tradeoff, where overly simple learning models underfit the data (suboptimal performance on the training data) and overly complex models are expected to overfit or \emph{memorize} the data (perfect training set performance, but suboptimal test set performance). The latter direction of this tradeoff has come into question with the observation that deep neural networks do not memorize their training data despite having sufficient capacity to do so \citep{zhang2016understanding}, the explanation of which is a matter of much interest.
	
	Due to their convenient gradient properties and excellent performance in practice, rectified-linear units (ReLU) have been widely adopted and are now ubiquitous in the field of deep learning. In addition, the relative simplicity of this function ($\max(\cdot, 0)$) makes the analysis of ReLU networks more straight-forward than networks with other activation functions.
	
	We propose a new notion of `non-linearity' of a ReLU layer with respect to an input batch. 
	We show that networks that networks that generalize well have deep layers that are approximately linear with respect to batches of similar inputs. In contrast,	networks that memorize their training data are highly non-linear with respect to similar inputs, even in deep layers.  
	
	Our method is based on the fact that the main source of non-linearity in ReLU networks is the threshold at zero. This thresholding determines the \emph{support} of the resulting activation matrix, which plays an important role in the analysis of non-negative matrices. As we discuss in Section \ref{sec:approach}, the \emph{non-negative rank} of a matrix is constrained by the shape of the support, and is therefore indicative of the degree of non-linearity in a ReLU activation matrix with respect to the input.
	
	Although computing the non-negative rank is NP-hard \citep{vavasis2009complexity}, we can restrict it with approximate \emph{non-negative matrix factorization} (NMF) \citep{lee1999learning}. 
	Consequently, we propose to estimate the `non-linearity' of a ReLU layer with respect to an input batch by performing NMF on a grid over the approximation rank $k$, and measuring the impact on network performance.
	
	This procedure can be seen as measuring the robustness of a neural network to increasing compression of its activations. We therefore compare our NMF-based approach to two additional dimensionality reduction techniques, namely principal component analysis (PCA) and random ablations.
	
	We report results for a variety of neural network architectures trained on several image and audio datasets. We show that our NMF-based approach is extremely sensitive to memorization in neural networks. We conduct a layer-by-layer analysis and our results reveal interesting details on the internal mechanism of memorization in neural networks. Finally, as an indicator for memorization, we use our proposed measure to perform early stopping.

	\section{Related Work}
	
	The study of factors involved in the bias-variance tradeoff in learning models goes back several decades.
	Classical results in statistical learning consider properties of learning models such as the VC-dimension~\citep{vapnik1998statistical} and Rademacher complexity~\citep{bartlett2002rademacher}. These properties give generalization bounds in terms of the capacity model to (over)fit data.
	When considering the vast capacity of deep neural networks, such bounds become irrelevant and fail to explain their ability to generalize well in practice \citep{zhang2016understanding,bartlett2017spectrally}.
	
	More direct analyses have been done with respect to a specific setting of model parameters. For instance, \citet{bartlett1998sample} showed that the number of weights in a network is less important compared to their scalar value (e.g. $\ell_2$-norm), and more recently \citet{bartlett2017spectrally} presented a bound for deep neural networks based on the product of spectral norms of the network's weight matrices. \citet{achille2017emergence} showed that memorizing networks contain more information in their weights.
	
	Methods to explain generalization have been proposed that examine a network's robustness to perturbations \citep{hochreiter1997flat,chaudhari2016entropy,keskar2016large,neyshabur2017exploring,li2017visualizing}. These methods propose the notion of \emph{flatness of minima} on the loss surface, assuming that perturbing the parameters without dramatically changing performance is an indicator of the generalization of a network.
	
	However, any reversible transformation, such as simple scaling, can arbitrarily manipulate the local flatness without affecting generalization \citep{dinh2017sharp}. The procedure we propose can be viewed as applying perturbations, albeit to \emph{activations} and not parameters, and must address this concern. The perturbations we apply to activations account for magnitude, since they depend on a change of rank or non-negative rank of the activation matrix, a property which is robust to rescaling and similar reversible transformations.
	
	In contrast to the methods described thus far, which deal exclusively with the parameters of the model, methods have been developed that account for the role of the data distribution. \citet{liang2017fisher} proposed to use the Fisher-Rao norm, which uses the geometry of the data distribution to weigh the contribution of different model parameters.
	The empirical studies of \citet{morcos2018importance} and \citet{novak2018sensitivity} explore robustness to specific types of noise. The former uses Gaussian noise and masking noise injected into hidden activations, while the latter interpolates between input samples to study network behavior on and off the data manifold. In both cases, robustness to noise proved a reliable indicator for good generalization. Additionally, \citet{arora2018stronger} derive generalization bounds in terms of robustness to noise.
	
	Our experimental setup is reminiscent of \citet{morcos2018importance} in that both methods apply a form of compression to hidden activations and test for robustness to this type of noise. Specifically, they set random axis-aligned directions in feature space to zero which can be viewed as a crude form of dimensionality reduction, i.e., by simply removing canonical dimensions. In our experiments we refer to this method as random ablations. Our results show that robustness to NMF compression is much more correlated with low memorization/high generalization than robustness to random ablations. \citet{arpit2017closer} have also studied various empirical aspect of memorization.
	
	As a dimensionality reduction technique, NMF has gained popularity due to its producing meaningful factorizations that lend themselves to qualitative interpretation across various domains such as document clustering \citep{xu2003document}, audio source separation~\citep{grais2011single}, and face recognition \citep{guillamet2002non}.
	In the context of deep convolutional neural networks, \citet{collins2018} applied NMF to the activations of an image classifier and showed that the result gives a decomposition into semantic parts, which benefits from the transformation invariance learned by the neural network.

	\section{Method} \label{sec:approach}
	
	Consider a ReLU layer of a neural network, parameterized by a weight matrix $\mW\in\sR^{m\times q}$. For a batch of $n$ inputs $\mX\in\sR^{n\times m}$, we compute the layer activation matrix $\mA$ as follows:
	
	\begin{align}
	\mA = \max\left(\mX\mW, 0\right) \in \sR_+^{n\times q}, \label{eq:activation}
	\end{align}
	
	where $\sR_+$ are the non-negative reals. We omit the bias term for notational convenience.
	
	\subsection{ReLU-induced support}
	
	\begin{figure}[t]
		\centering
		\bgroup
		\setlength{\tabcolsep}{0pt}
		\begin{tabularx}{\textwidth}{rYYY}
			& (a) Linear & (b) Approximately linear &  (c) Highly non-linear \\
			& $rc(M)=1$ & $rc(M)=3$ &  $rc(M)=8$ \\
			\rotatebox[origin=c]{90}{\centering Samples}   &
			\raisebox{-.5\height}{\includegraphics[width=.27\textwidth]{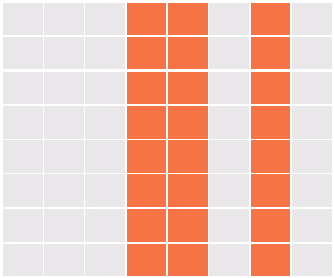}}&
			\raisebox{-.5\height}{\includegraphics[width=.27\textwidth]{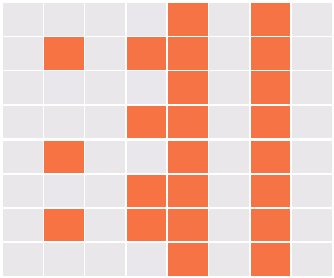}}&
			\raisebox{-.5\height}{\includegraphics[width=.27\textwidth]{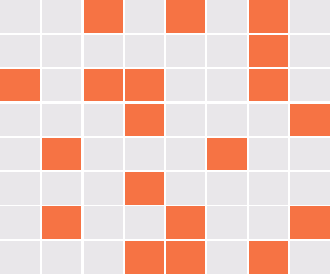}}  \vspace{2pt}\\
			& \multicolumn{3}{c}{Channels} \\
		\end{tabularx}
		\egroup
		\caption{
			\textbf{The support of the activation matrix} is determined by ReLU threshold. (a) When all the rows of the support are identical, there is a sub-weight-matrix such that the layer is fully linear with respect to the input batch. (b, c) As the support becomes more complex, which we characterize by the increase in its rectangle cover number, the layer becomes more non-linear.
		}
		\label{fig:supp}
	\end{figure}
	
	The processing of a single input $\vx$ by a ReLU network is equivalent to sampling a sub-network that is linear with respect to the sample \citep{wang2016analysis}. This could be accomplished by simply setting to zero the columns of each $\mW$ whose dot product with the input is negative (and would thus be set to zero by ReLU), and then removing the thresholding.\footnote{This is a similar intuition to that of viewing dropout as an approximation to a model ensemble, where the dropout mask is seen to sample a sub-network \citep{srivastava2014dropout}.}
	
	Extending this notion to a batch of several input samples to a ReLU layer, suppose the samples are sufficiently close to each other such that they all share the same ReLU mask $\vm\in\{0,1\}^q$. In this case, we may say that \emph{the layer is linear with respect to its input batch}. This is because, for the entire batch, instead of using ReLU, we could zero out a subset of columns and have have a linear system, i.e., $\mA = \mX\mW\text{diag}(\vm)$.
	
	For an activation matrix $\mA$ (Equation~\ref{eq:activation}), we consider the support $\mM=\text{supp}(\mA)$, which we describe as a binary 0/1 matrix where $\mM_{i,j} = 1$ where $\mA_{i,j}>0$. Because $\mA$ is a ReLU activation matrix, the structure of $\mM$ is mainly determined by the thresholding at zero.\footnote{The probability of an activation value being exactly zero prior to thresholding is negligible.} Because threshold is the main source of non-linearity in a ReLU network, the support takes on a special meaning in this case.

	\subsection{Rectangle cover number and non-negative rank} \label{sec:Rectangle cover number and non-negative rank}
	We want to characterize how close to being linear a layer is with respect to its input $\mX$ by examining the support of the resulting activations $\mM$. If all the rows of $\mM$ are identical to a unique vector $\vm$, we can say the layer is completely linear with respect to $\mX$. In general, the `simpler' the support $\mM$, the closer to linearity the layer.
	
	One measure that captures this idea is the \emph{rectangle cover number} of a matrix, $rc(\mM)$, an important quantity in the study of communication complexity \citep{klauck2003rectangle}. Also known as the \emph{Boolean rank}, $rc(\mM)$ is the smallest number $r$ for which there exist binary matrices $\mU_B\in\{0,1\}^{n\times r}$, $\mV_B\in\{0,1\}^{r\times q}$ such that their \emph{Boolean} matrix multiplication satisfies $\mM = \mU_B\mV_B$. As a complexity measure for ReLU activations, $rc(\mM)=1$ means the layer is linear with respect to its input, and higher values $rc(\mM)$ imply increasing non-linearity. This is visualized in Figure~\ref{fig:supp}.
	
	
	Intuitively, imagine having to fit a layer with `ReLU switches', each of which controls a subset of weight matrix columns. In the linear case, one switch would control all the columns. In the most non-linear case we would require a switch for every column, which is also the maximal value of $rc(\mM)$.
	
	Because computing the rectangle cover number $rc(\mM)$ is complex, several approximations and bounds to it have been studied \citep{fiorini2013combinatorial}. For the support of a non-negative matrix, a well-known upper-bound is:
	
	\begin{align}
	rc(\text{supp}(\mA))\leq rank_+(\mA) ,
	\end{align}
	
	where $rank_+(A)$ is the \emph{non-negative rank} of $\mA$  \citep{gillis2012geometric} that is defined as the smallest number $k$ such that there exist non-negative matrices $\mU_+\in\sR_+^{n\times k}$, $\mV_+\in\sR_+^{k\times q}$ such that $A = U_+V_+$. Similar to the rectangle cover number, the non-negative rank is hard-constrained by the combinatorial arrangement of $\text{supp}(\mA)$, but additionally accounts for the actual value in the non-zero entries of $\mA$.
	
	While computing $rank_+(\mA)$ is not easier than computing $rc(\text{supp}(\mA))$, we can \emph{restrict it} by performing approximate non-negative matrix factorization (NMF).
	
	\subsection{Non-negative matrix factorization}
	
	For a given non-negative rank constraint $k$, NMF solves for:
	
	\begin{align}
	\argmin_{\mU_+,\mV_+} \|\mA - \mU_+\mV_+\|_2^2, \label{eq:nmf}
	\end{align}
	
	with $\mU_+,\mV_+$ as defined above. The result $\mU_+\mV_+ = \tilde{\mA}_k \approx \mA$ is the closest matrix to $\mA$ under the Frobenius norm that has $rank_+$ at most $k$.
	
	Consequently, we propose to estimate the `linearity' of a ReLU layer with respect to a batch of similar inputs by performing NMF on a grid over the non-negative rank $k$, and measuring the impact on network performance by observing the change in the prediction (output layer) as we change $k$. This procedure also addresses the fact that in practice network activations tend to be noisy, whereas $\text{supp}(\mA)$ is not robust to noise, i.e., $\mA_{i,j} = \epsilon > 0 \rightarrow \mM_{i,j}=1$ even for very small $\epsilon$.
	
	Concretely, if we let $\mA_i$ be the activation matrix at layer $i$, during the forward pass, we replace the feature activations of one or several layers with their rank $k$ NMF approximations:
	\begin{align}
	\mA_{i+1} = \max \left( \tilde{\mA}_kW_{i+1}, 0 \right)
	\label{eq:fwnmf}
	\end{align}
	
	For convolutional networks, we first reshape the tensor of feature maps from $n\times q\times h\times w$ to $(n\cdot h\cdot w)\times q$, i.e., we flatten the batch ($n$) and spatial dimensions ($h,w$) to form a matrix with $q$ columns, where $q$ is the number of channels in that layer.
	We then inversely reshape the approximated features to continue forward propagation through the network.
	
	\subsection{Single class batches}
	
	We now characterize the input batch, with respect to which we would like to measure layer linearity.
	Informally, the goal of training is to cluster together input samples that have similar (or identical) output labels,while separating them from samples of other labels. In the context of classification then, we expect therefore that from a certain depth and onward, samples of the same class will have similar activations, and thus a simpler support.
	In other words, while a network may exploit flexible non-linear structure to separate \emph{different classes}, we expect that with respect to a \emph{single class}, deep layers are approximately linear.
	
	
	When single-class batches are \emph{not} approximately linear, even in deep layers, we take this as evidence of memorization, i.e., the samples are not tightly clustered. In the next section we present empirical results that support this view.

	\section{Experiments} \label{sec:experiments}

	\subsection{Feature compression and memorization}
	We start by studying networks that have been forced into different level of memorization due to label randomization on their training set \citep{zhang2016understanding}. The level of induced memorization is controlled by setting a probability $p$ for a training label to be randomized, i.e., $p=0$ is the unmodified dataset and $p=1$ gives fully random labels. Note that the capacity of these networks is sufficiently large such that the training accuracy is 1 in all cases, regardless of the value of $p$.
	
	We perform experimental evaluations on several image datasets, namely CIFAR-10~\citep{krizhevsky2009learning}, Fashion-MNIST~\citep{xiao2017fashion}, SVHN\citep{netzer2011reading}, and ImageNet~\citep{ILSVRC15}, as well as on the Urban Sounds audio classification dataset \citep{salamon2014urban}. We use a fully-connected network for Fashion-MNIST and various CNN architectures for the others, which we describe in more detail the appendix.
	
	\begin{figure}[t]
		\centering
		\bgroup
		\setlength{\tabcolsep}{0pt}
		\begin{tabularx}{\textwidth}{lYYlY}
			& (a) {\tt conv2\_1} & (b) {\tt conv4\_1} &  & (c) All layers \\
			\rotatebox[origin=c]{90}{\centering Accuracy}   &
			\raisebox{-.5\height}{\includegraphics[width=.3\textwidth]{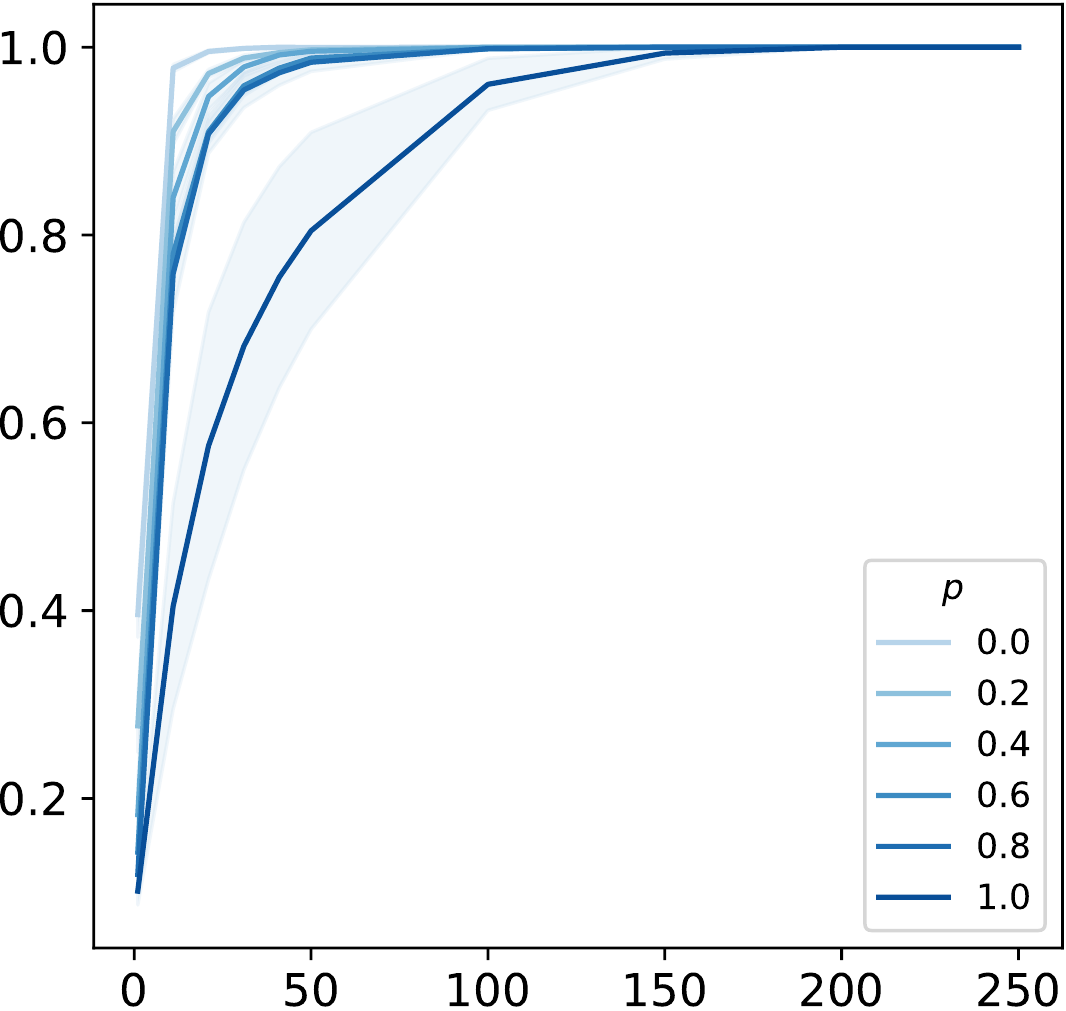}}&
			\raisebox{-.5\height}{\includegraphics[width=.3\textwidth]{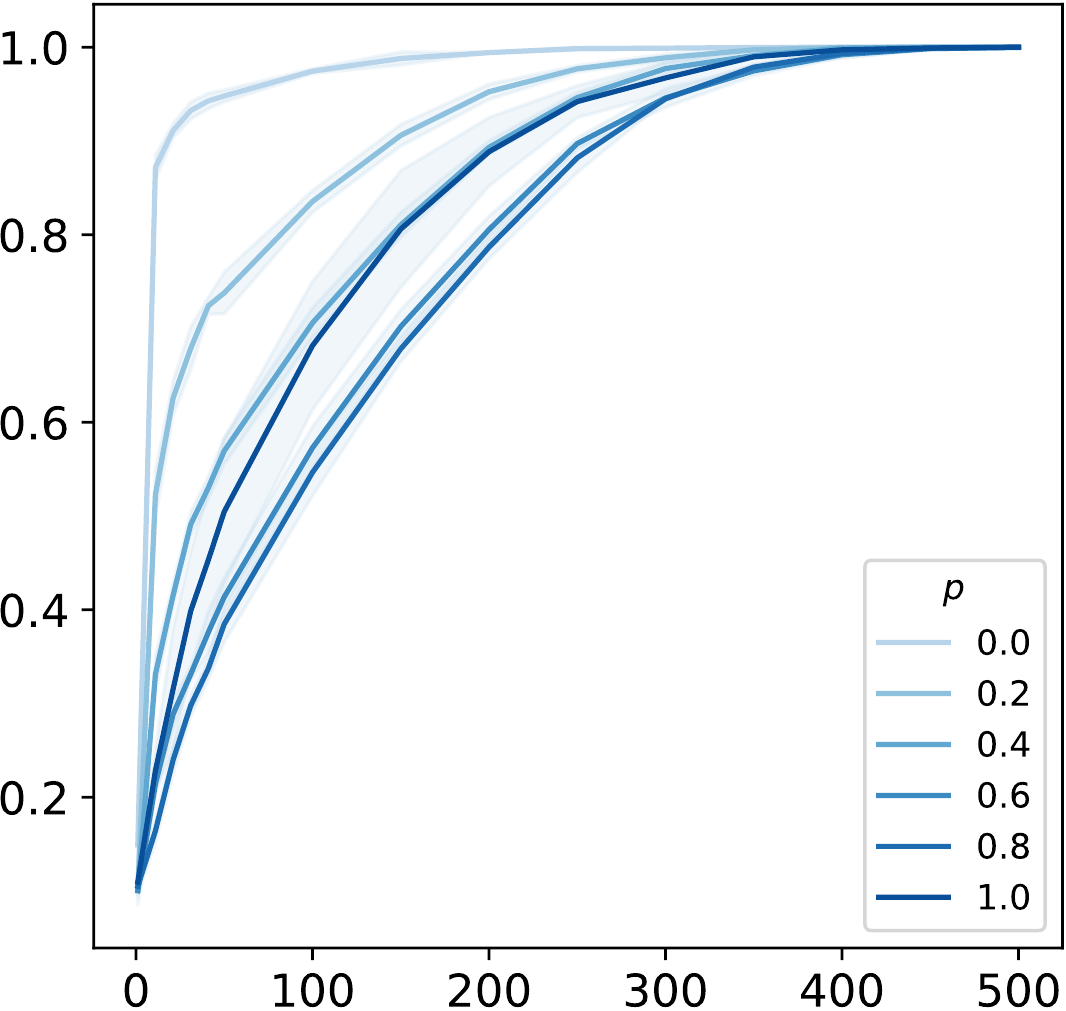}}&
			\rotatebox[origin=c]{90}{\centering AuC}   &
			\raisebox{-.5\height}{\includegraphics[width=.3\textwidth]{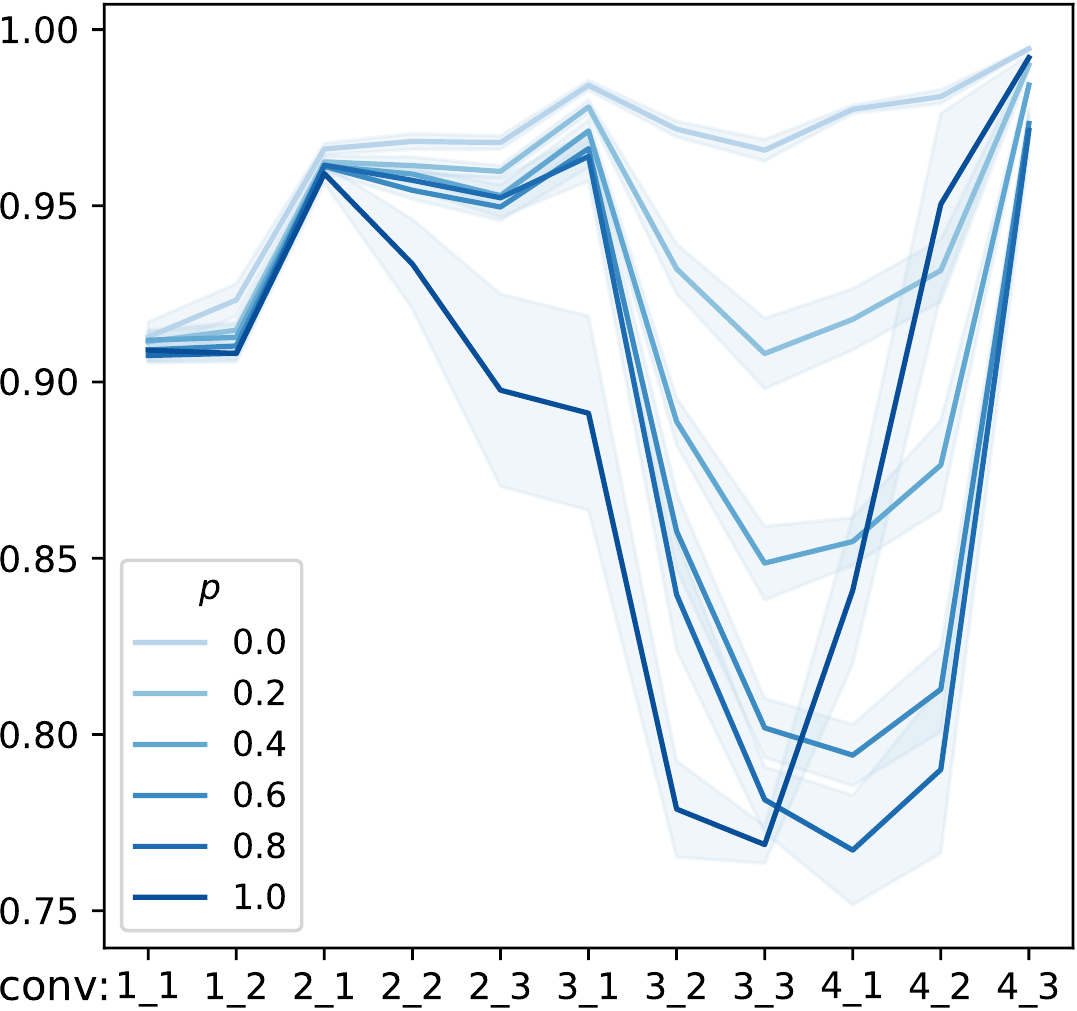}}  \vspace{2pt}\\
			& Non-negative rank $k$  & Non-negative rank $k$  & & Layer \\
		\end{tabularx}
		\egroup
		\caption{
			\textbf{Memorization mechanism} of CNNs trained on CIFAR-10, with increasing level of label randomization $p$ (i.e., $p=0$ is the unmodified dataset).
			We analyze each layer by applying NMF compression to its activation matrix with increasing rank $k$, while observing the impact on classification performance. In (a) and (b) we show the $k$ vs. accuracy curves at an early layer and a deep layer, respectively. We can immediately see that in deep layers, networks with high memorization are significantly less robust to NMF compression, indicating higher degrees of non-linearity. Furthermore, networks trained on fully randomized labels ($p=1$) behave differently than networks with partial or no randomization. By summarizing each curve in (a) and (b) by its area under the curve (AuC), we show in (c) a birds-eye view over all layers. All networks with $p<1$ pass through distinct phases consisting of a \emph{feature extraction} phase until {\tt conv3\_1}, followed by a \emph{memorization} phase until {\tt conv4\_2}, followed by a final \emph{clustering} phase. Interestingly, the case $p=1$ shifts the process into earlier layers, explaining why layer-by-layer it appears as an outlier.
		}
		\label{fig:CIFAR-lbl}
	\end{figure}

	\subsubsection{Layer by layer analysis}
	
	We start by analyzing the layers of an 11-layer CNN trained on CIFAR-10. We sampled 10 batches (one batch per class) of 50 images, and compressed the activation matrix at each layer individually down to various values of the non-negative rank. We then measured classification accuracy of the prediction. In this analysis we report average results for 60 neural networks, ten networks (with different random initializations) trained per randomization level $p$.
	
	In Figure~\ref{fig:CIFAR-lbl} (a) and (b) we show $k$ vs. accuracy curves of networks trained with increasing levels of label randomization, at an early layer ({\tt conv2\_1}) and a deep layer ({\tt conv4\_1}) respectively. We can immediately see that networks trained on fully randomized labels ($p=1$) behave differently than networks with partial or no randomization. Furthermore, note that in deep layers, memorizing networks are significantly less robust to NMF compression, i.e., their activations posses a high non-negative rank, which indicates high non-linearity with respect to the input, as discussed in Section \ref{sec:Rectangle cover number and non-negative rank}. We can characterize each curve in (a) and (b) with a single number, its area under the curve (AuC). This allows us in Figure~\ref{fig:CIFAR-lbl} (c) to generate a single figure for all layers. Networks with $p<1$ display a similar feed-forward trend up until layer {\tt conv3\_1}. Since these networks differ from each other in no way other than the level of label randomization on the training data, we take this to be a generic \emph{feature extraction} phase common to all of them. In the next phase, until {\tt conv4\_2}, we see a big difference between networks, such that more memorization (higher $p$) is correlated with lower AuC, i.e., higher non-negative rank and hence non-linearity of those layers with respect to single-class batches. We therefore localize memorization to these layers. Lastly, the  phase only of {\tt conv4\_3} is where samples of the same class are clustered together, right before the final 10-dimensional classification layer (which is not shown). This final phase is in accordance with the premise that regardless of randomization level $p$, all of these networks achieve perfect training set accuracy.Interestingly, setting $p=1$ shifts the process to earlier layers, explaining why layer-by-layer this case appears as an outlier.

	\subsubsection{Compression techniques}
	\label{sec:feat_mem_net}
	
	We compare different compression techniques with regards to detecting memorization. Given our choice of solving NMF under the Frobenius norm (Equation~\ref{eq:nmf}), a natural method to compare against is principal component analysis (PCA), which optimizes under the same norm but without the non-negativity constraint on its factors. We also consider random ablations, i.e., setting a random subset of columns in the activation matrix to zero, since this technique has been used previously to detect memorization~\citep{morcos2018importance}.
	
	Rather than choosing a single layer, we sequentially apply compression to several layers. We target the final convolutional blocks of our CNNs, all of which contain three layers, each of which consists of 512 channels. In fully-connected networks, we applied compression to all layers.
	
\begin{figure}
		\centering
		\bgroup
		\setlength{\tabcolsep}{0.1pt}
		\begin{tabular}{cccc}
			& (a) Single-class NMF & (b) Single-class PCA &  (c) Random Ablations \\
			\rotatebox[origin=c]{90}{\centering Accuracy} \hspace{1.0pt} &
			\raisebox{-.5\height}{\includegraphics[width=.320\textwidth]{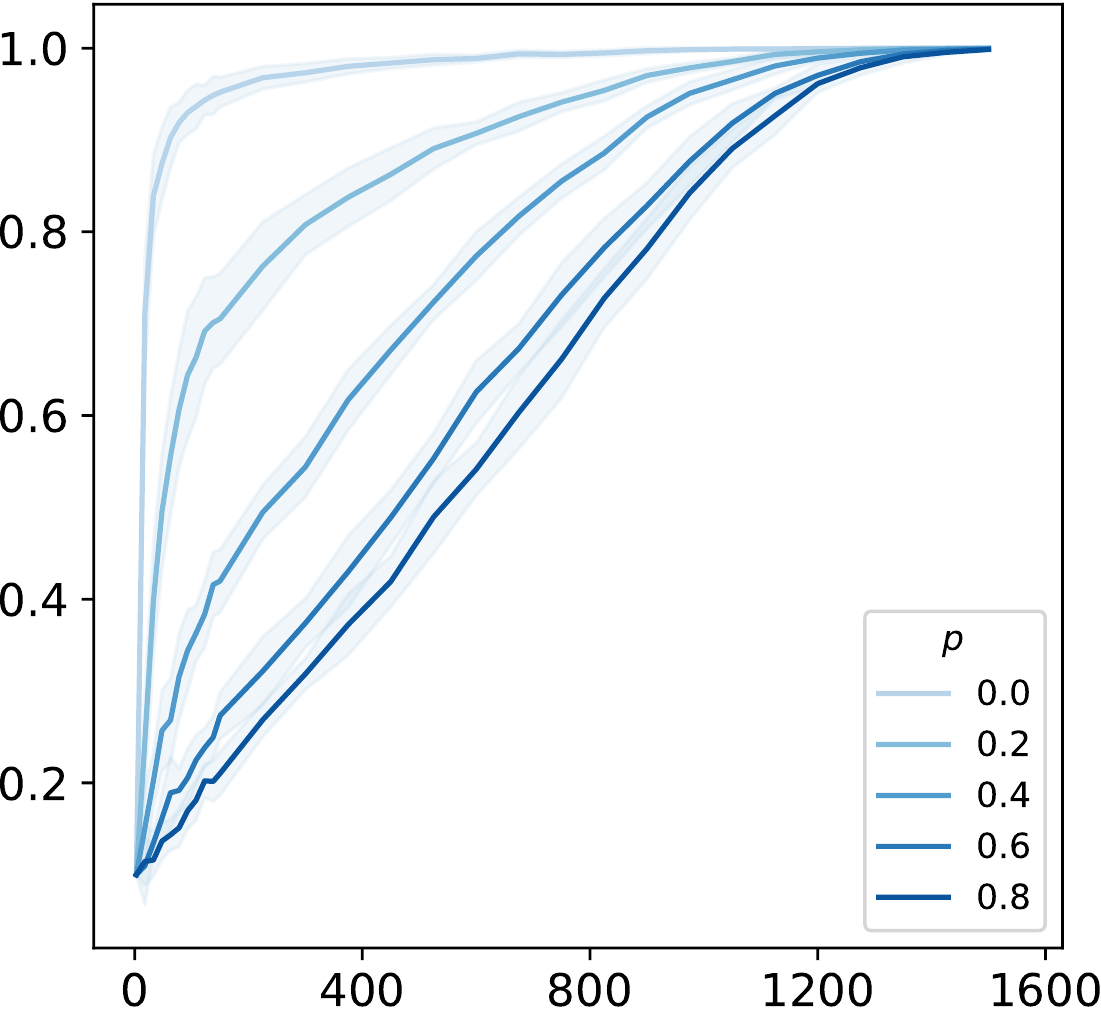}}&
			\raisebox{-.5\height}{\includegraphics[width=.320\textwidth]{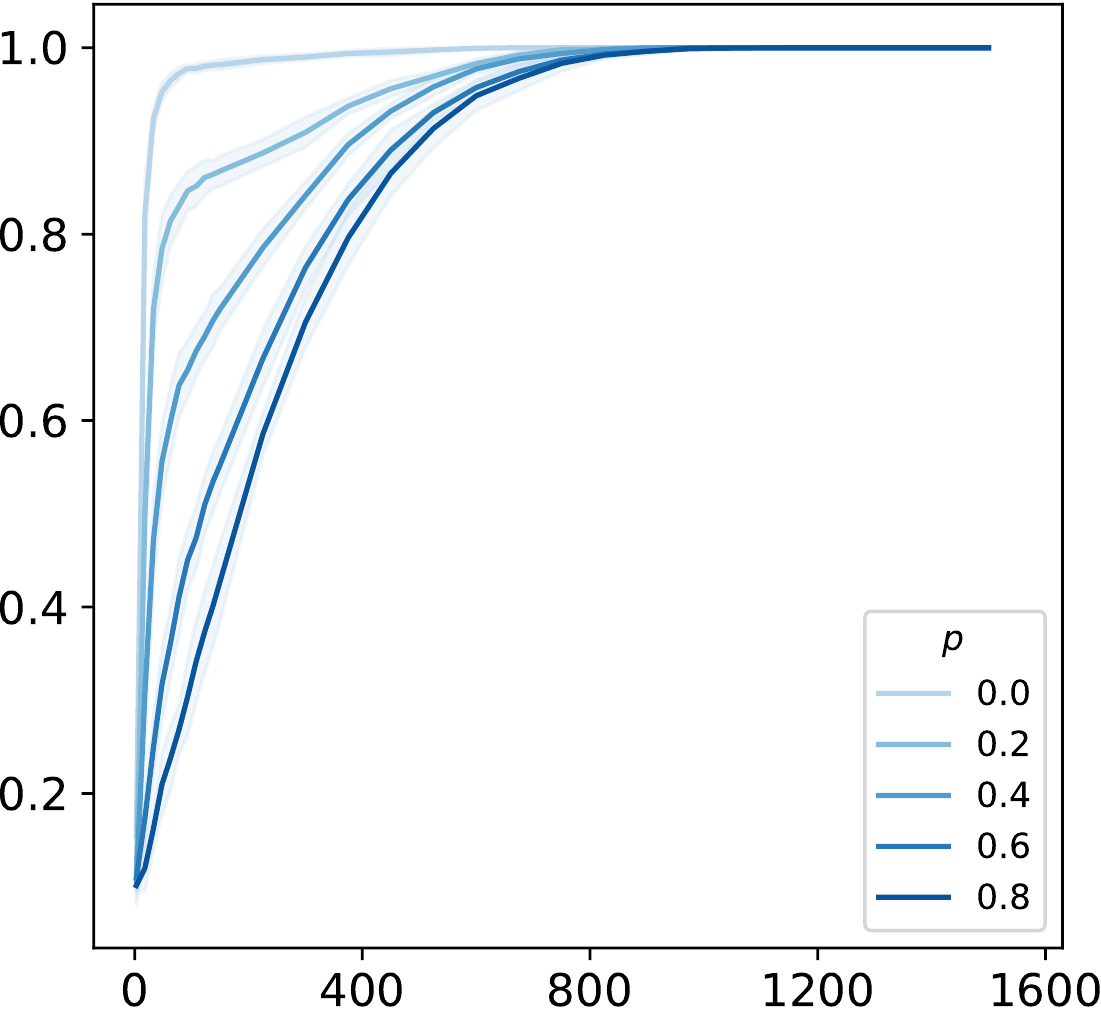}}&
			\raisebox{-.5\height}{\includegraphics[width=.320\textwidth]{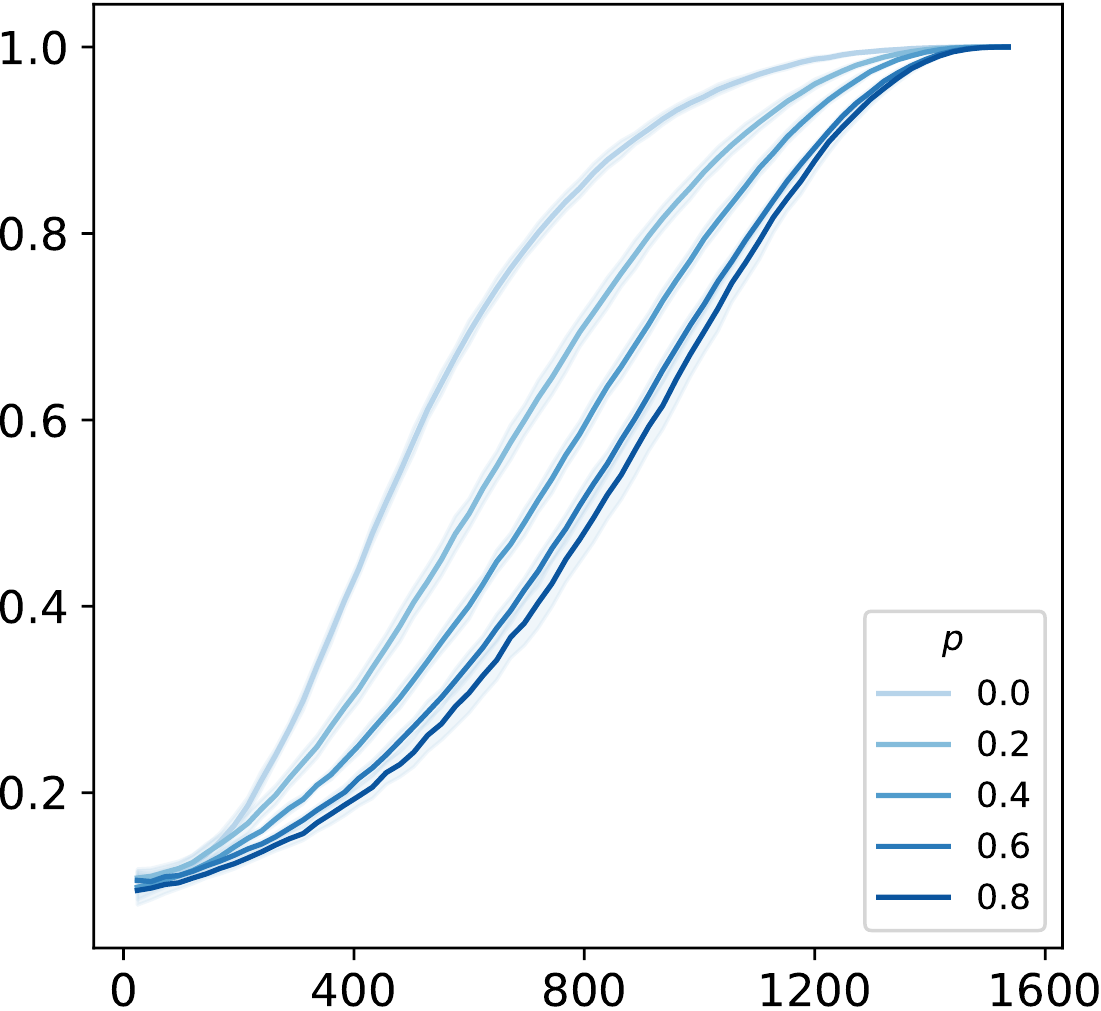}}  \vspace{2pt}\\
			& Non-negative rank $k$ & Rank $k$ & Rank $k$\\
		\end{tabular}
		\egroup
		\caption{
			\textbf{Detecting memorization via compression}. We demonstrate this on networks trained with different levels of label randomization ($p$), and hence of memorization. (a) Due to its sensitivity to the non-linearity of ReLU activations, NMF successfully captures the level of memorization present in neural networks.
			(b) PCA compression is able to regain sufficient variance for good accuracy even with small values of $k$, which renders it less effective for detecting memorization.
			(c) Though not taking into account batch statistics, random ablations can distinguish between different levels of memorization, albeit less significantly.
			Compression was applied to the final three (convolutional) layers of CNNs trained on CIFAR-10.
		}
		\label{fig:CIFAR}
	\end{figure}

	In Figure \ref{fig:CIFAR} we give results for the CIFAR-10 dataset, which confirm that NMF compression is indeed more sensitive to memorization, due to the properties of the non-negative rank discussed in Section \ref{sec:Rectangle cover number and non-negative rank}. PCA, which is less constrained, is more `efficient' at compressing the activations, but is in turn less discriminative with respect to the level of memorization. Finally, we confirm that robustness to random ablations correlates with less memorization, however less so than NMF.
	
\begin{figure}[t]
		\bgroup
		\setlength{\tabcolsep}{0.9pt}
		\begin{tabular}{cccc}
			& Fahsion-MNIST & SVHN &  Urban Sounds \\
			& (a) Single-class NMF & (b) Single-class NMF & (c) Single-class NMF \\
			\rotatebox[origin=c]{90}{\centering Accuracy} \hspace{1.0pt}&
			\raisebox{-.5\height}{\includegraphics[width=.315\textwidth]{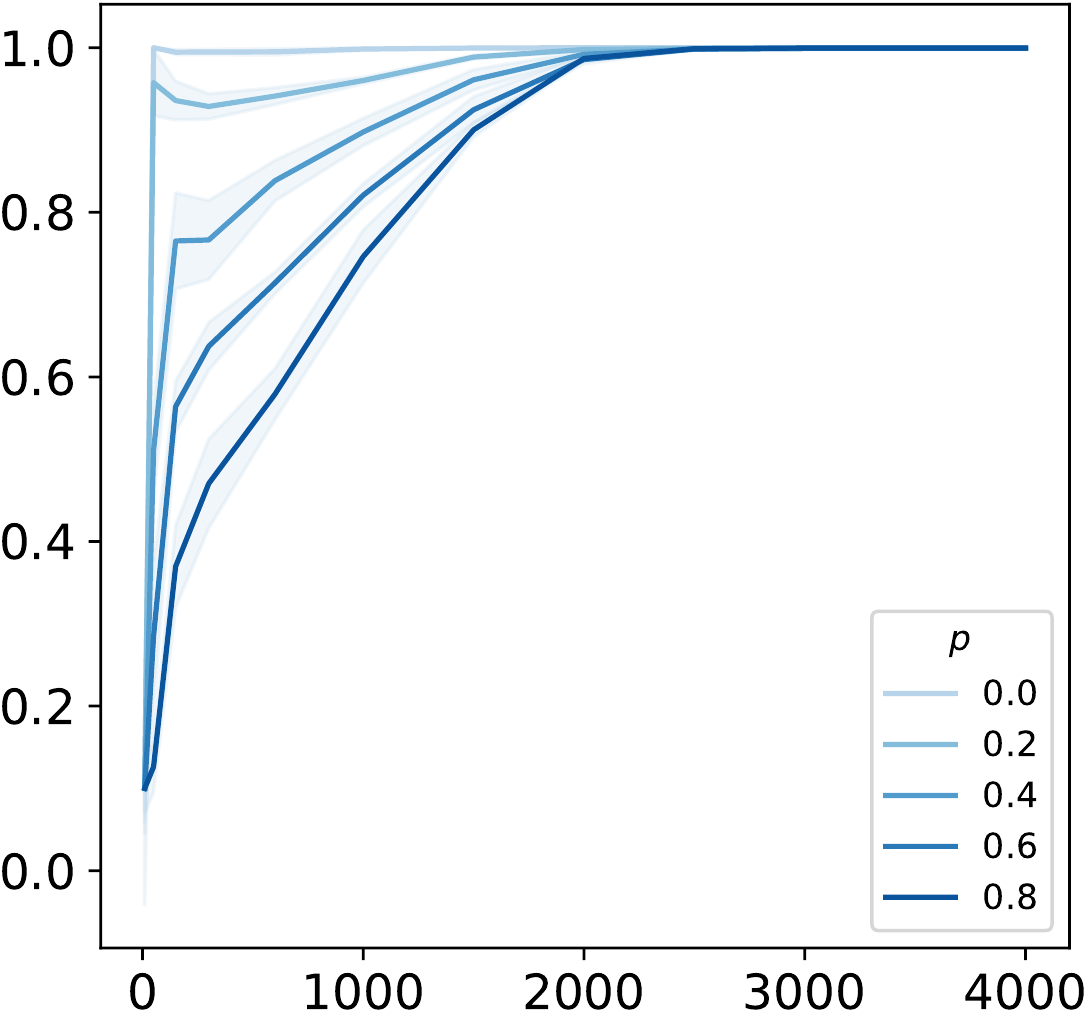}}&
			\raisebox{-.5\height}{\includegraphics[width=.315\textwidth]{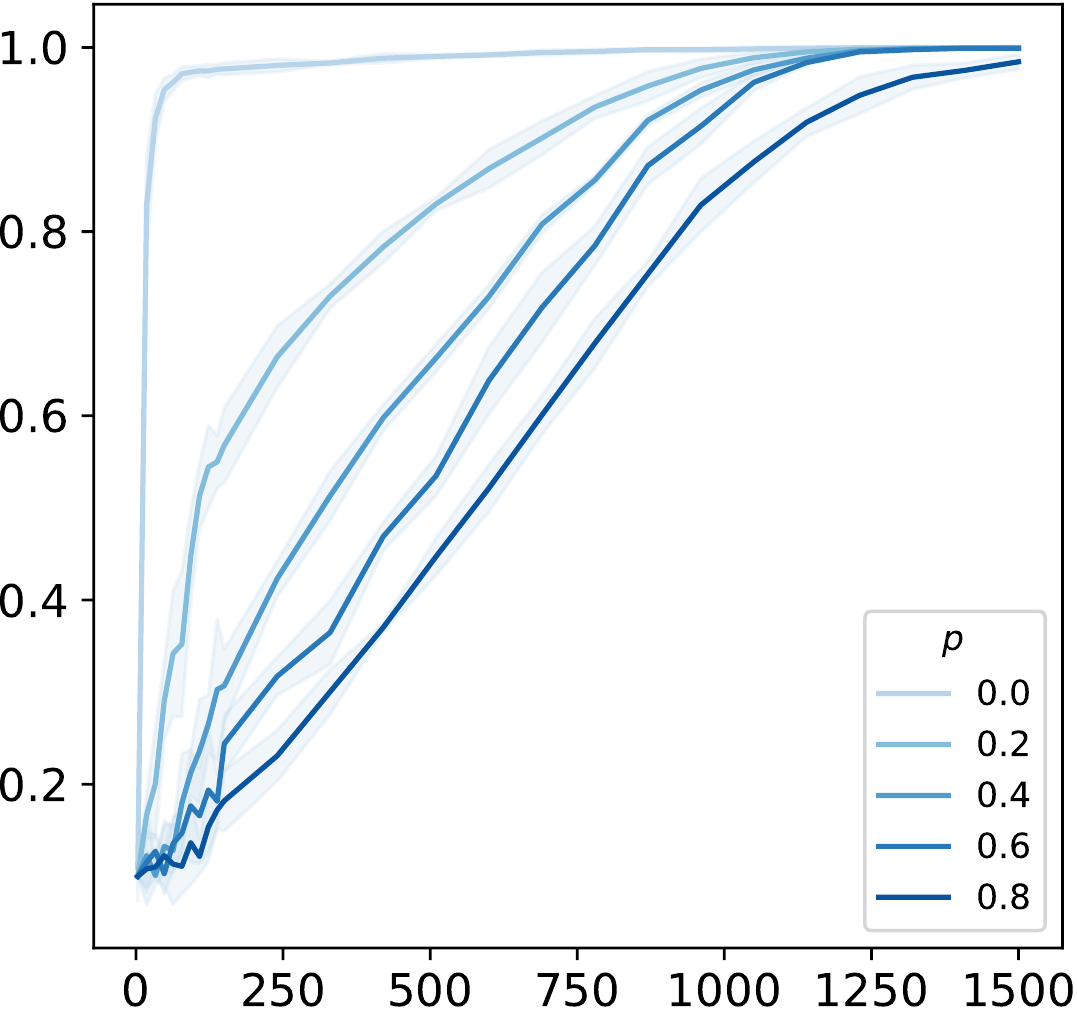}}&
			\raisebox{-.5\height}{\includegraphics[width=.315\textwidth]{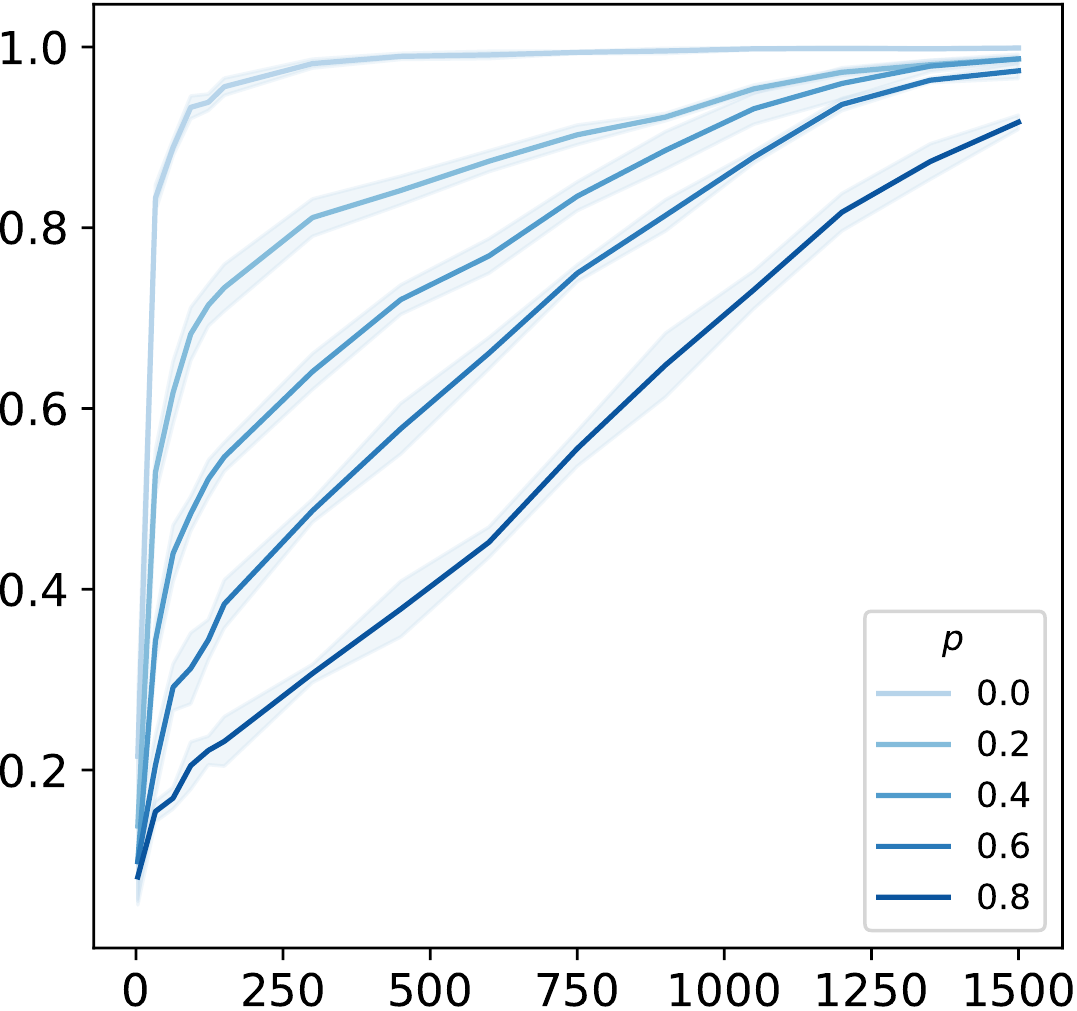}}  \vspace{2pt}\\
			& \multicolumn{3}{c}{Non-negative rank $k$}\\
			\\
			& (d) Multi-class NMF & (e) Multi-class PCA  & (f) Random ablations \\
			\rotatebox[origin=c]{90}{\centering Accuracy} \hspace{1.0pt} &
			\raisebox{-.5\height}{\includegraphics[width=.315\textwidth]{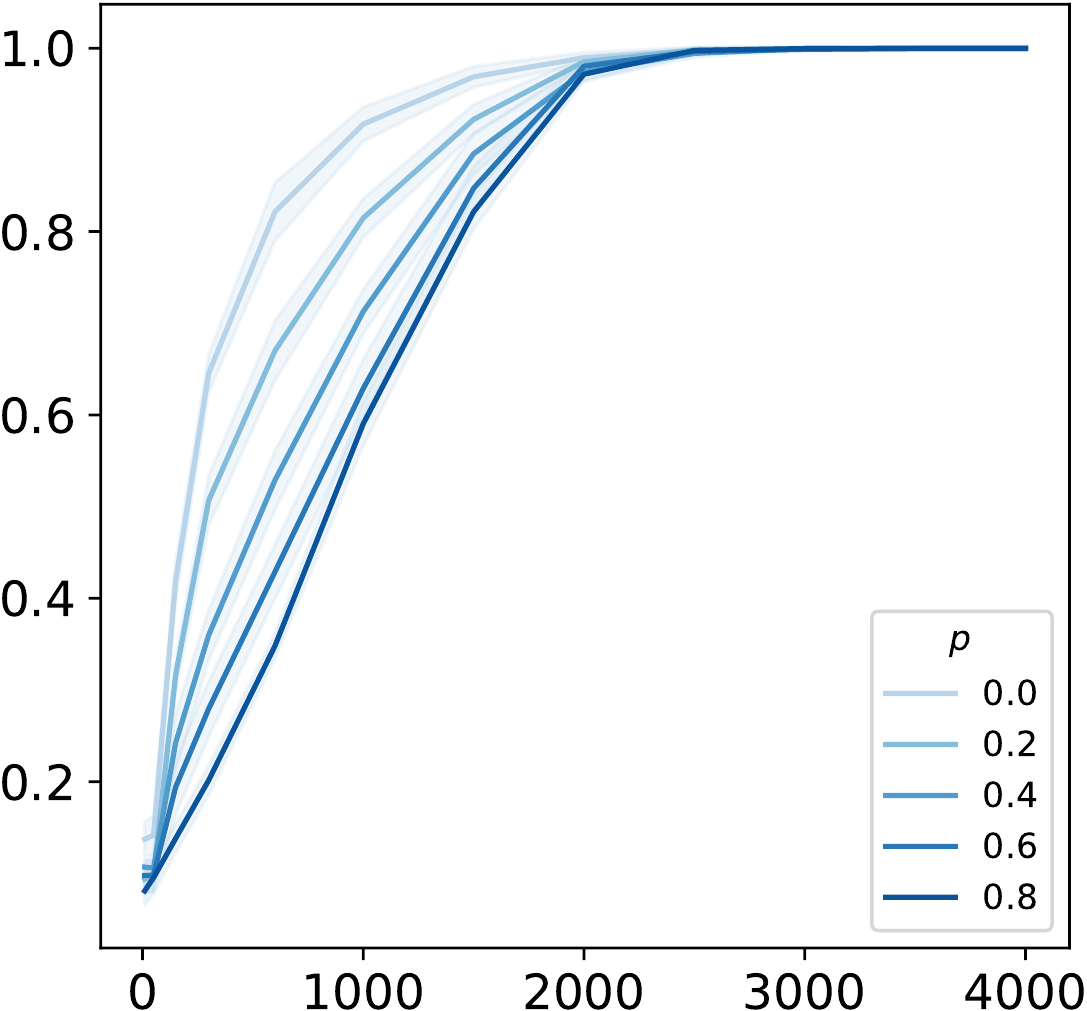}}&
			\raisebox{-.5\height}{\includegraphics[width=.315\textwidth]{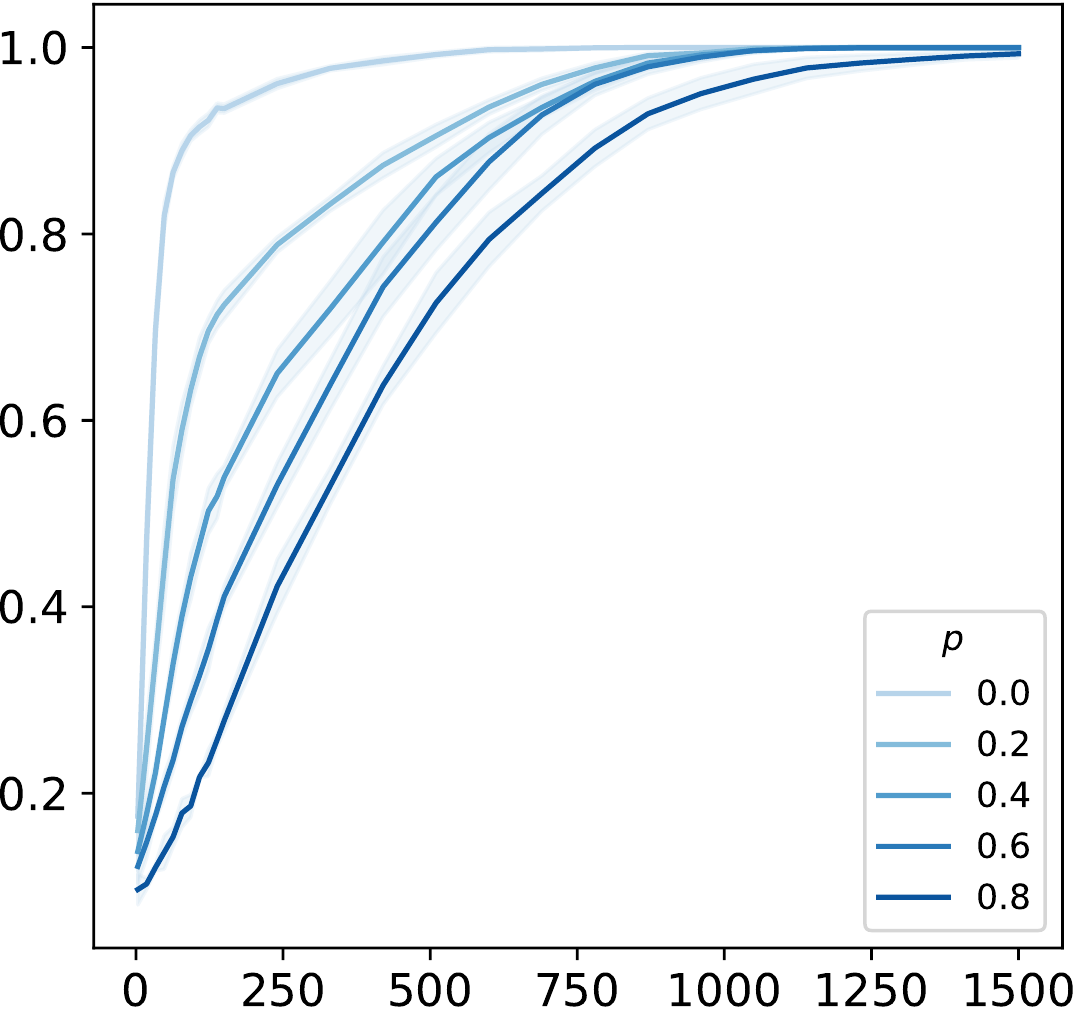}}&
			\raisebox{-.5\height}{\includegraphics[width=.315\textwidth]{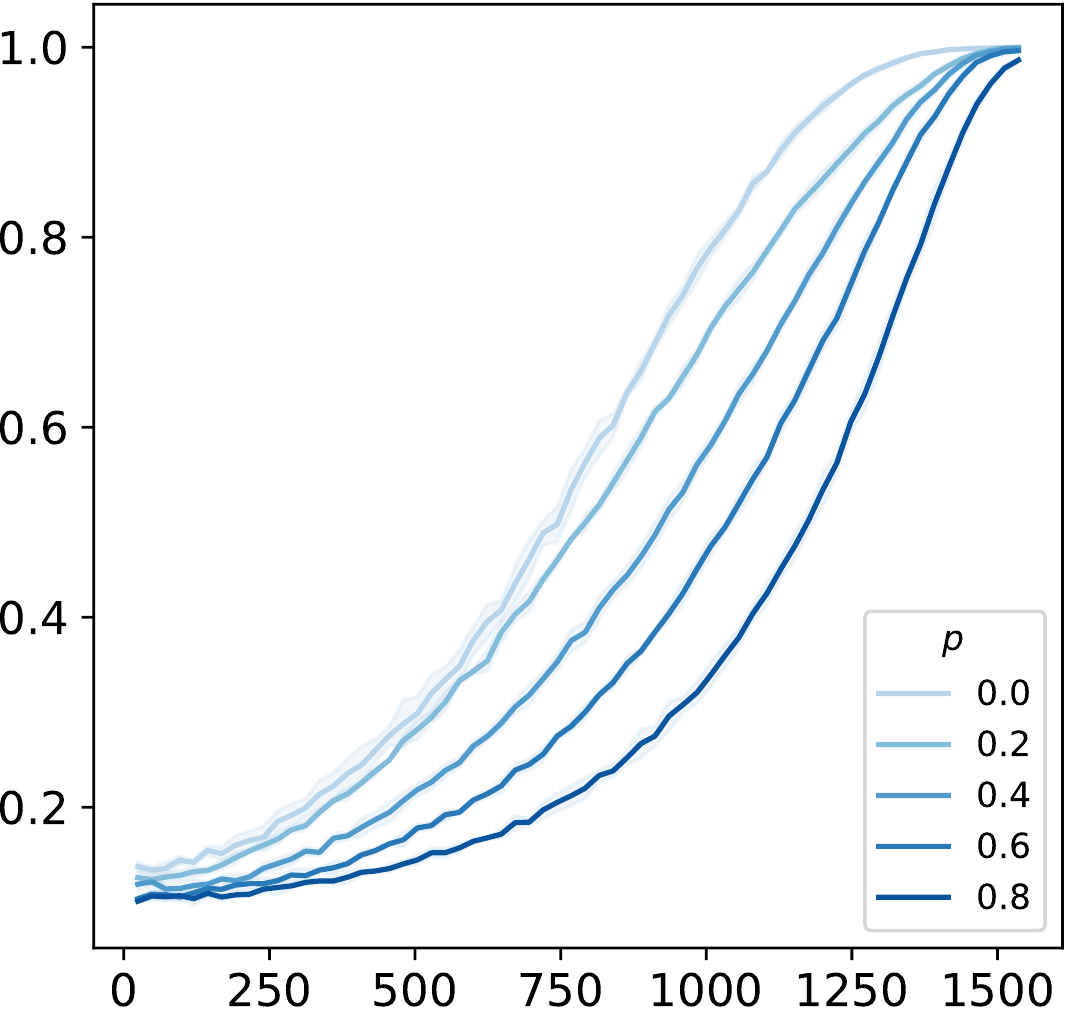}}  \vspace{2pt} \\
			& Non-negative rank $k$ & Rank $k$ & Rank $k$\\
		\end{tabular}
		\egroup
		\caption{
			\textbf{Memorization across various datasets and network architectures}.
			We show that NMF-based compression is sensitive to memorization in diverse settings. Each column shows results for a
			specific dataset and network architecture. (a, b, d, e) We show that network layers are considerably
			more linear with respect to single-class batched than with respect to multi-class batches.  (b,c,e,f)
			PCA and random ablations show less sensitivity to memorization compared with NMF.}
		\label{fig:mnist}
	\end{figure}

	In Figure \ref{fig:mnist} we show additional results for single-class NMF on three additional datasets and network architectures (described in the appendix), including a fully-connected network for Fashion-MNIST. The results in (d) and (e), of applying PCA and NMF to multi-class batches, show that such batches produce activations with higher rank or non-negative rank compared to single-class batches. This is a result of the network trying to separate samples of different labels.

	\subsection{Feature compression and generalization}
	We have shown results for networks forced into memorization due to label randomization. 
	In this section we show our technique is useful for predicting good generalization in a more realistic setting, without artificial noise.
	
	In addition to the experiments below, we refer the reader to Section \ref{sec:vgg} in the appendix, where predict \emph{per-class} generalization of a pre-trained VGG-19 network on a ImageNet classes.
	
	\subsubsection{Compression techniques}
\begin{figure}[t]
		\centering
		\bgroup
		\setlength{\tabcolsep}{0.1pt}
		\begin{tabular}{cccc}
			& (a) Single-class NMF & (b) Single-class PCA &  (c) Random Ablations \\
			\rotatebox[origin=c]{90}{\centering Normalized AuC} \hspace{1.0pt} &
			\raisebox{-.5\height}{\includegraphics[width=.320\textwidth]{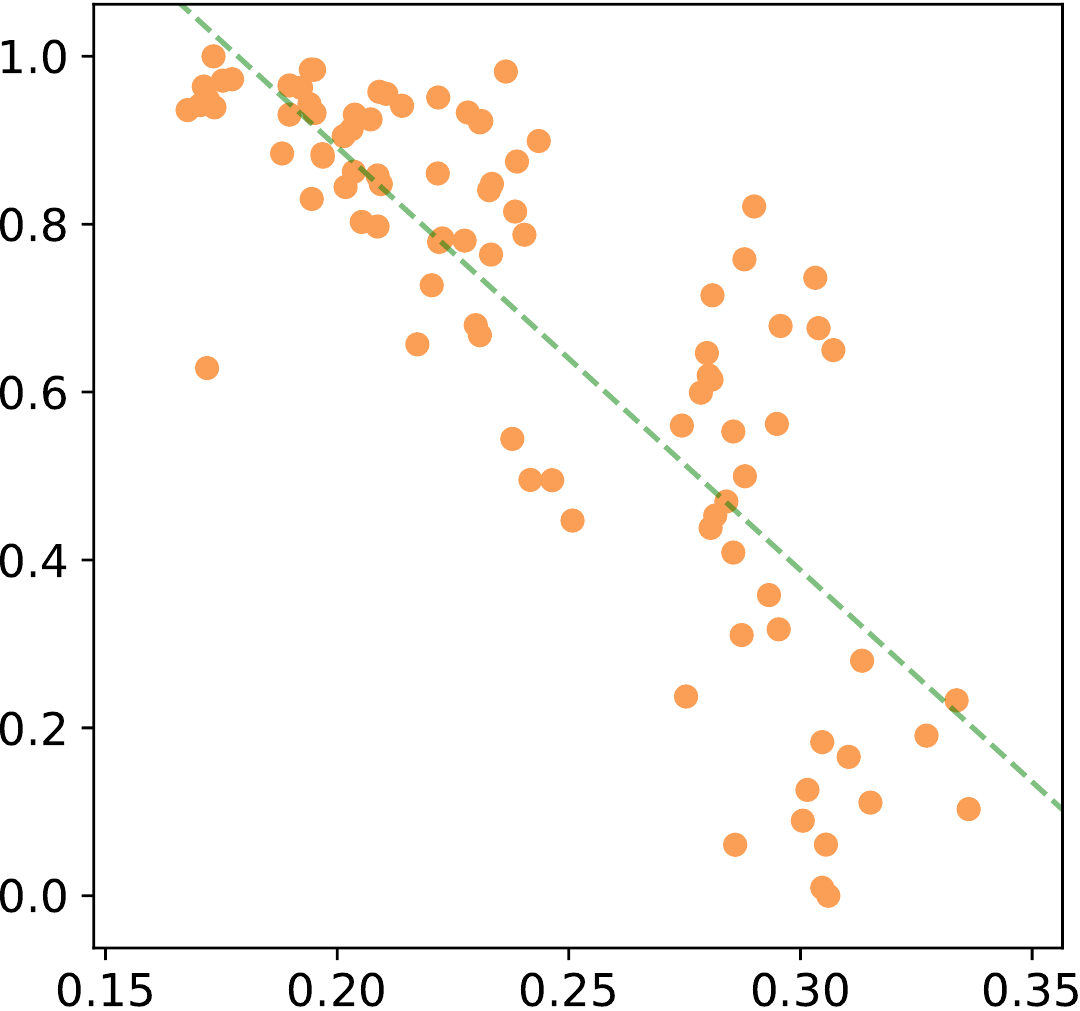}}&
			\raisebox{-.5\height}{\includegraphics[width=.320\textwidth]{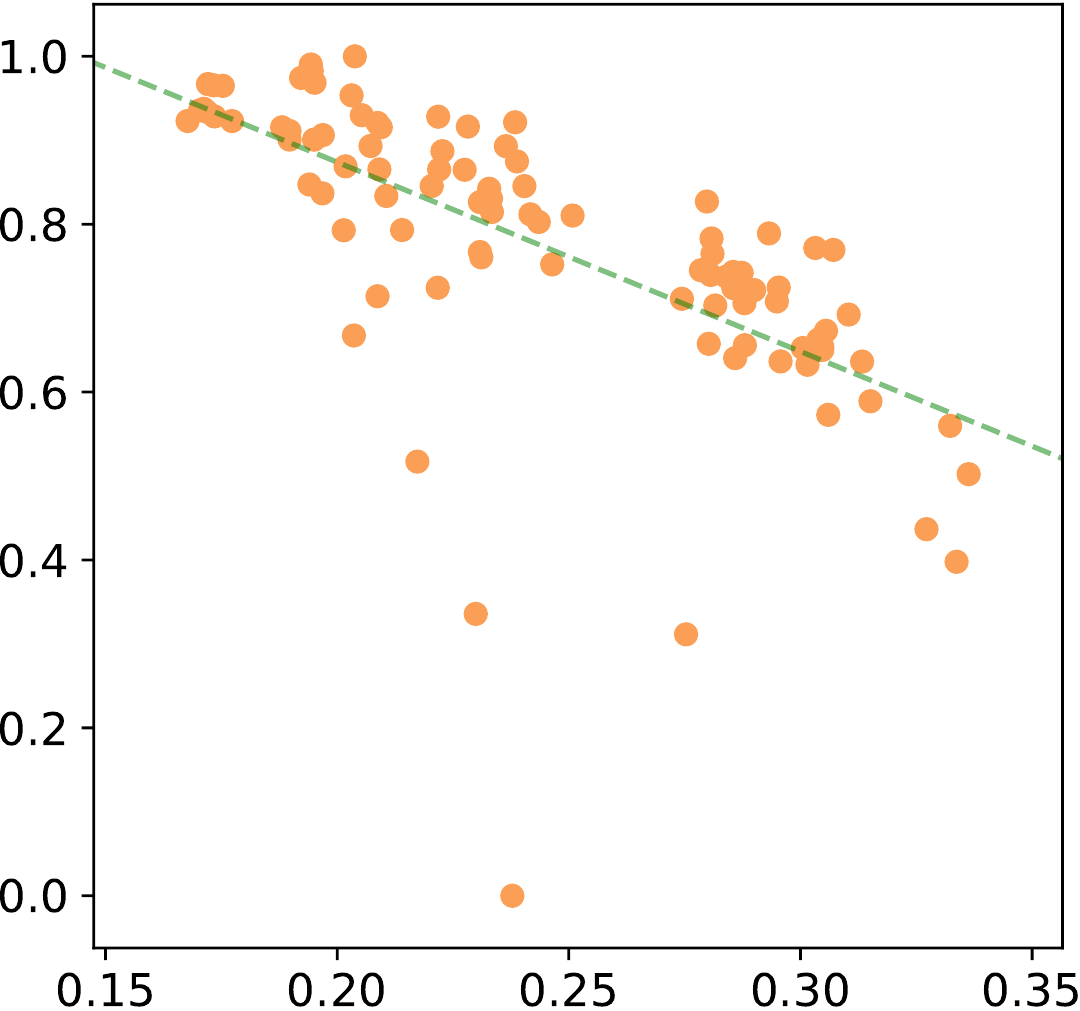}}&
			\raisebox{-.5\height}{\includegraphics[width=.320\textwidth]{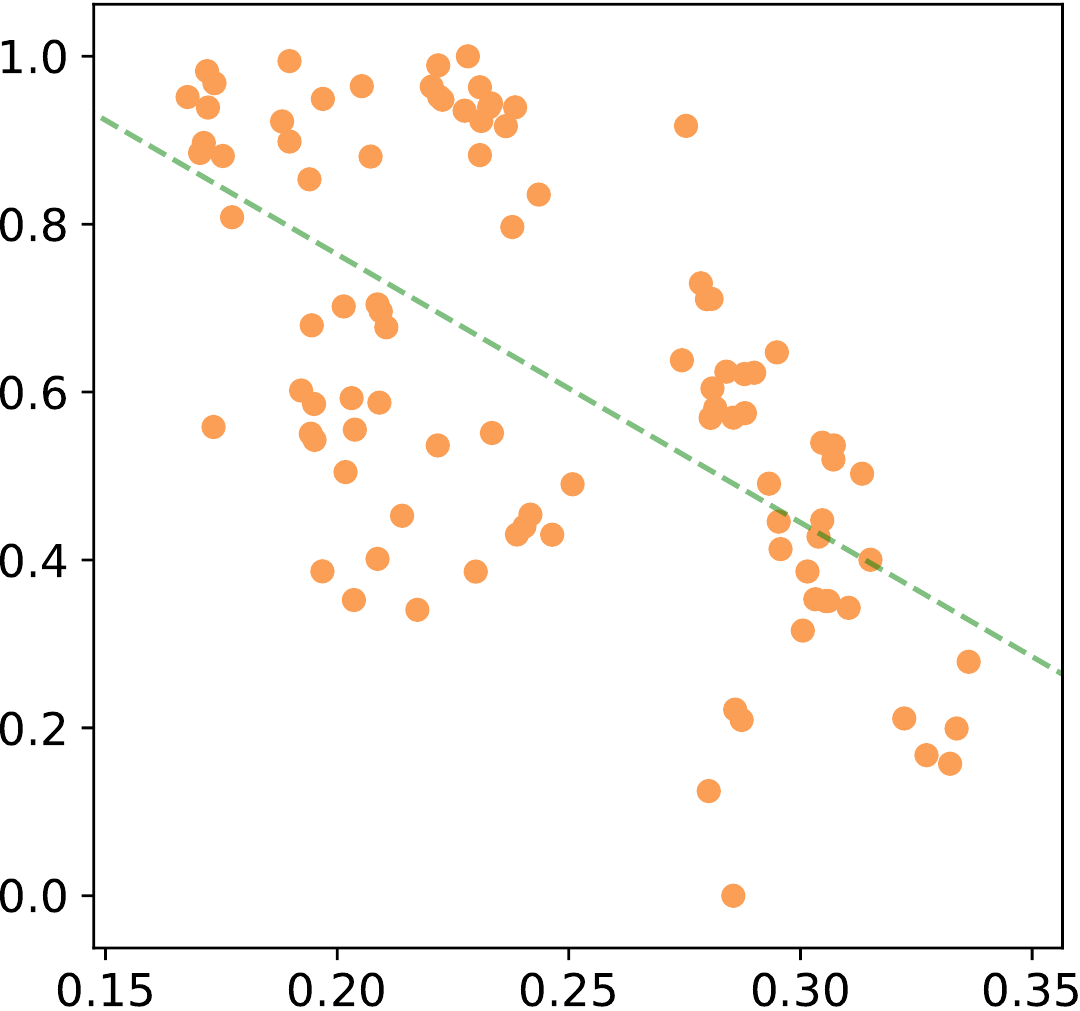}}  \vspace{2pt}\\
			& \multicolumn{3}{c}{Generalization error} \\
		\end{tabular}
		\egroup
		\caption{
			\textbf{Detecting generalization via compression}. While all three methods show correlation with generalization error, NMF is most correlated with a Pearson correlation of -0.82, followed by PCA with -.064 and random ablation with -0.61.
		}
		\label{fig:generalization}
	\end{figure}

	We trained 96 CNN classifiers on CIFAR-10, over a grid of hyper-parameter values for the batch size, weight decay and optimization algorithm, SGD vs. ADAM \citep{kingma2015adam}. Following the same procedure as above, for each of the three methods, NMF, PCA, or random ablations, we computed the $k$ vs. accuracy curves for each network, targeting its final convolutional block. In Figure~\ref{fig:generalization} we compare the area under the curve (AuC) of each curve with the average generalization error on the test set.
	
	While all three methods show correlation with generalization error, NMF is most correlated with a Pearson correlation of -0.82, followed by PCA with -.064 and random ablation with -0.61.
	
	\subsubsection{Early stopping}
	
	We test whether our method can detect memorization during training.
	Doing so would allow us to perform early stopping, i.e., stop training as memorization begins to decrease generalization.
	
	We trained CNNs on CIFAR-10 with the original labels.
	Each network was trained for 10K batches with a batch size of 100. We recorded the test set error every 250 batches,
	and tested the non-linearity of the deepest three convolutional layers using our NMF-based approach with a coarse grid on $k$.
	As before, we compute the area under each $k$ vs. accuracy curve as in Figures~\ref{fig:CIFAR} (c).
	Finally, we also computed the area under the curve produced by random ablations.
	
\begin{figure}
		\bgroup
		\setlength{\tabcolsep}{0.5pt}
		\begin{tabular}{cccc}
			\rotatebox[origin=c]{90}{\centering Test loss}&
			\raisebox{-.5\height}{\includegraphics[width=0.32\textwidth]{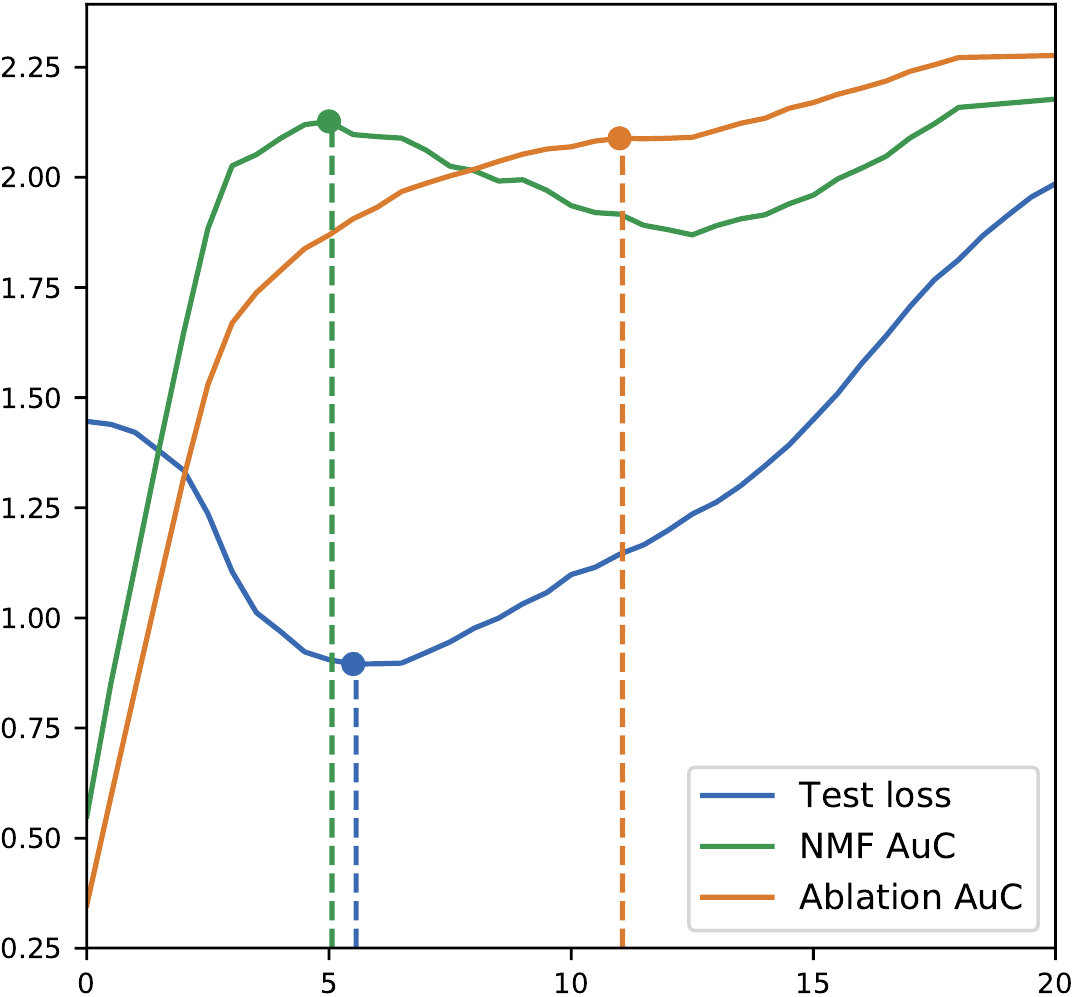}} &
			\raisebox{-.5\height}{\includegraphics[width=0.32\textwidth]{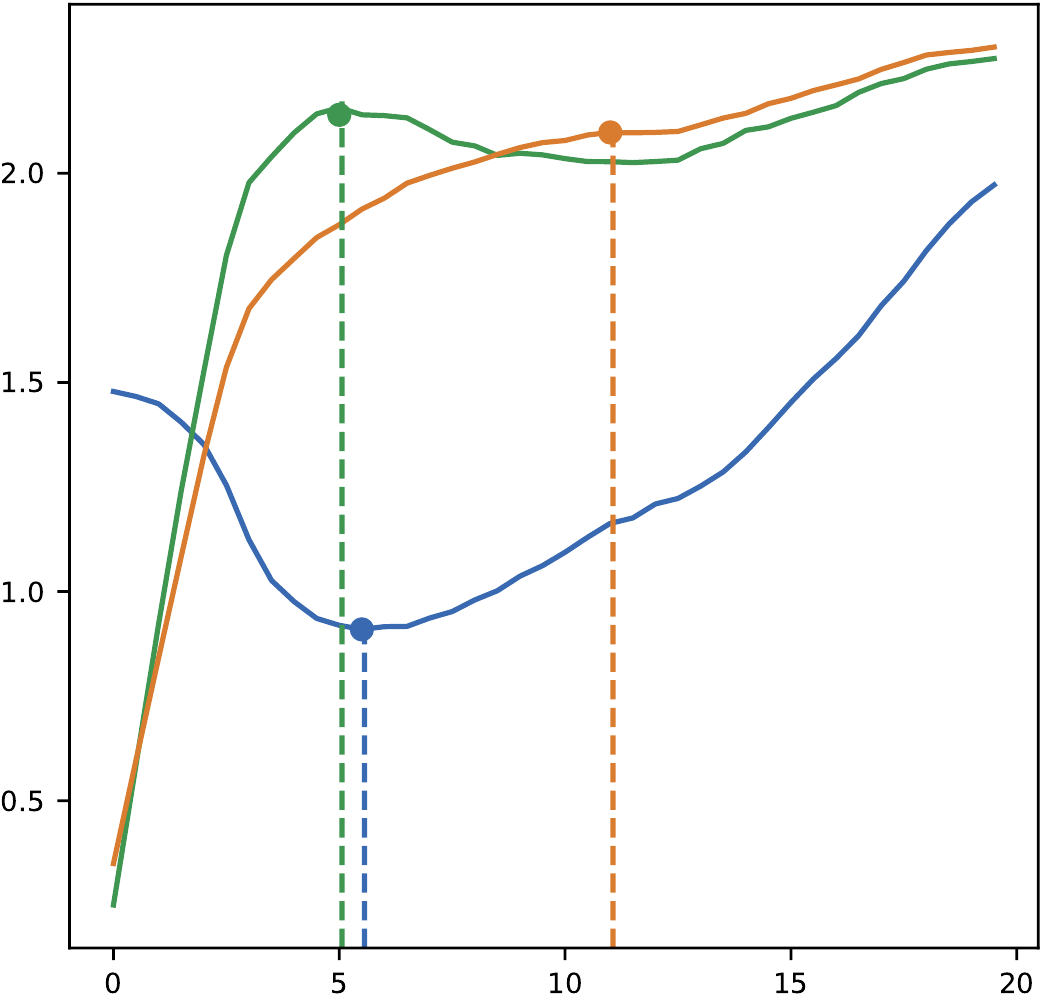}} &
			\raisebox{-.5\height}{\includegraphics[width=0.32\textwidth]{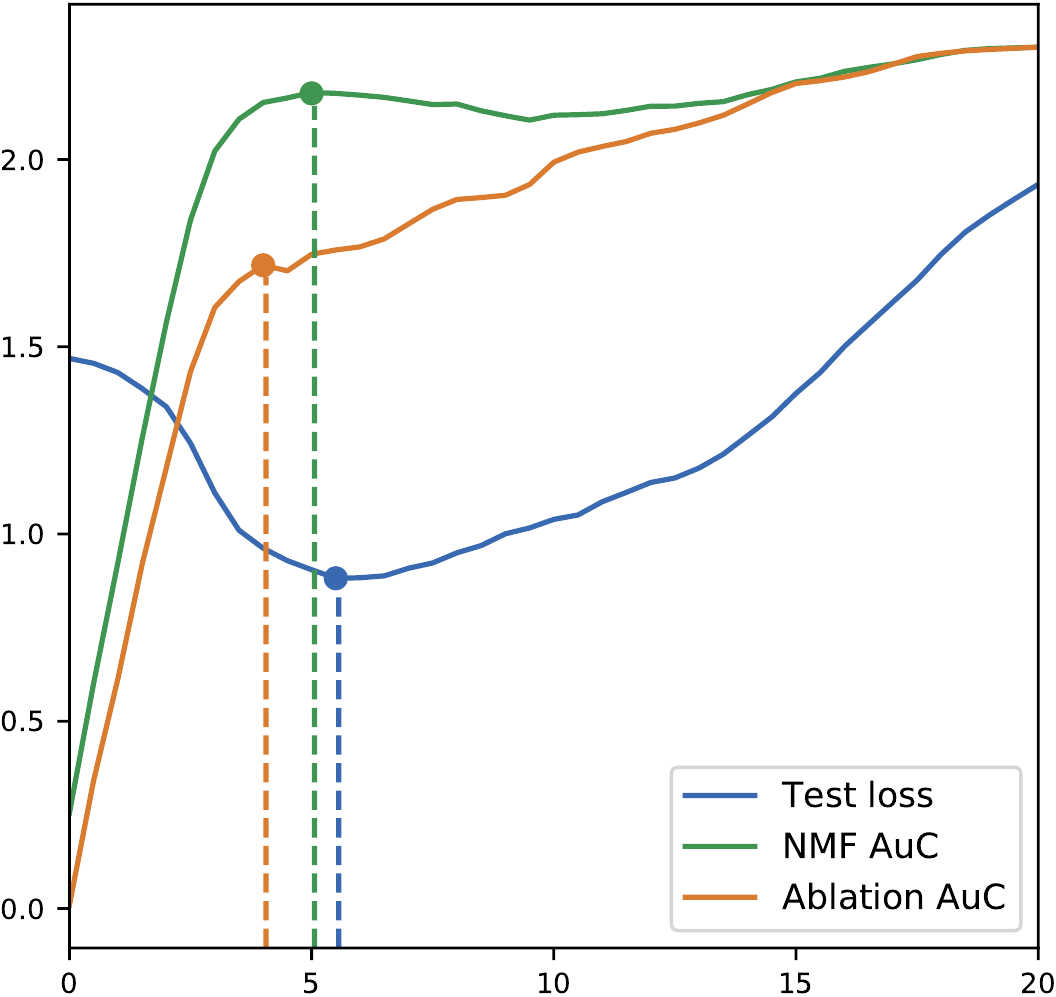}} \vspace{2pt} \\
			& \multicolumn{3}{c}{Epoch}
		\end{tabular}
		\egroup
		\caption{
			\textbf{Early stopping} for CNN training on CIFAR-10.
			The test error is (in blue) starts to increase after about the 5th epochs, indicating the start of overfitting.
			Using our proposed single-class NMF approach, we can detect the test error turning point.
			We show the area under the curve (AuC) for the single-class NMF approach (in green) for the accuracy measures as discussed in Section~\ref{sec:feat_mem_net}.
			Similarly, we show the AuC when performing random ablations (in orange).
			The NMF AuC curve correlates with the test error curve, as seen by the proximity of their extrema.
		}
		\label{fig:cifar-earlystop}
	\end{figure}

	Results of three instances with different initializations are shown in Figure~\ref{fig:cifar-earlystop}.
	These include the test error (in blue), our single-class NMF method (in green), and random ablations (in orange).
	We smooth the plots using a radius of two epochs to reduce the noise.
	The blue dashed lines mark the local minima of the test loss in Figure~\ref{fig:cifar-earlystop}.
	The green and orange dashed lines indicate the location of the first local maxima of the NMF and random ablation AuC curves after smoothing has been applied.
	We notice that the test loss minima align almost precisely with the maximum NMF AuC.

	\section{Conclusion}
	
	We have introduced a notion of a ReLU layer's non-linearity with respect to an input batch, which is based on its proximity to a linear system. We measure this property indirectly via NMF applied to deep activations of single-class batches. While more analysis is required before definite guarantees could be given, we find that our approach is successful in detecting memorization and generalization across a variety of neural network architectures and datasets.
	
	\clearpage
	
	\bibliography{references}

\begin{thebibliography}{35}
\providecommand{\natexlab}[1]{#1}
\providecommand{\url}[1]{\texttt{#1}}
\expandafter\ifx\csname urlstyle\endcsname\relax
  \providecommand{\doi}[1]{doi: #1}\else
  \providecommand{\doi}{doi: \begingroup \urlstyle{rm}\Url}\fi

\bibitem[Achille \& Soatto(2017)Achille and Soatto]{achille2017emergence}
Alessandro Achille and Stefano Soatto.
\newblock Emergence of invariance and disentangling in deep representations.
\newblock \emph{arXiv preprint arXiv:1706.01350}, 2017.

\bibitem[Arora et~al.(2018)Arora, Ge, Neyshabur, and Zhang]{arora2018stronger}
Sanjeev Arora, Rong Ge, Behnam Neyshabur, and Yi~Zhang.
\newblock Stronger generalization bounds for deep nets via a compression
  approach.
\newblock \emph{arXiv preprint arXiv:1802.05296}, 2018.

\bibitem[Arpit et~al.(2017)Arpit, Jastrz{\k{e}}bski, Ballas, Krueger, Bengio,
  Kanwal, Maharaj, Fischer, Courville, Bengio, et~al.]{arpit2017closer}
Devansh Arpit, Stanis{\l}aw Jastrz{\k{e}}bski, Nicolas Ballas, David Krueger,
  Emmanuel Bengio, Maxinder~S Kanwal, Tegan Maharaj, Asja Fischer, Aaron
  Courville, Yoshua Bengio, et~al.
\newblock A closer look at memorization in deep networks.
\newblock \emph{International Conference on Machine Learning (ICML)}, 2017.

\bibitem[Bartlett(1998)]{bartlett1998sample}
Peter~L Bartlett.
\newblock The sample complexity of pattern classification with neural networks:
  the size of the weights is more important than the size of the network.
\newblock \emph{IEEE transactions on Information Theory}, 44\penalty0
  (2):\penalty0 525--536, 1998.

\bibitem[Bartlett \& Mendelson(2002)Bartlett and
  Mendelson]{bartlett2002rademacher}
Peter~L Bartlett and Shahar Mendelson.
\newblock Rademacher and gaussian complexities: Risk bounds and structural
  results.
\newblock \emph{Journal of Machine Learning Research}, 3\penalty0
  (Nov):\penalty0 463--482, 2002.

\bibitem[Bartlett et~al.(2017)Bartlett, Foster, and
  Telgarsky]{bartlett2017spectrally}
Peter~L Bartlett, Dylan~J Foster, and Matus~J Telgarsky.
\newblock Spectrally-normalized margin bounds for neural networks.
\newblock In \emph{Advances in Neural Information Processing Systems}, pp.\
  6240--6249, 2017.

\bibitem[Chaudhari et~al.(2016)Chaudhari, Choromanska, Soatto, LeCun, Baldassi,
  Borgs, Chayes, Sagun, and Zecchina]{chaudhari2016entropy}
Pratik Chaudhari, Anna Choromanska, Stefano Soatto, Yann LeCun, Carlo Baldassi,
  Christian Borgs, Jennifer Chayes, Levent Sagun, and Riccardo Zecchina.
\newblock Entropy-sgd: Biasing gradient descent into wide valleys.
\newblock \emph{arXiv preprint arXiv:1611.01838}, 2016.

\bibitem[Collins et~al.(2018)Collins, Achanta, and Susstrunk]{collins2018}
Edo Collins, Radhakrishna Achanta, and Sabine Susstrunk.
\newblock Deep feature factorization for concept discovery.
\newblock In \emph{The European Conference on Computer Vision (ECCV)}, 2018.

\bibitem[Dinh et~al.(2017)Dinh, Pascanu, Bengio, and Bengio]{dinh2017sharp}
Laurent Dinh, Razvan Pascanu, Samy Bengio, and Yoshua Bengio.
\newblock Sharp minima can generalize for deep nets.
\newblock \emph{arXiv preprint arXiv:1703.04933}, 2017.

\bibitem[Fiorini et~al.(2013)Fiorini, Kaibel, Pashkovich, and
  Theis]{fiorini2013combinatorial}
Samuel Fiorini, Volker Kaibel, Kanstantsin Pashkovich, and Dirk~Oliver Theis.
\newblock Combinatorial bounds on nonnegative rank and extended formulations.
\newblock \emph{Discrete Mathematics}, 313\penalty0 (1):\penalty0 67 -- 83,
  2013.

\bibitem[Gillis \& Glineur(2012)Gillis and Glineur]{gillis2012geometric}
Nicolas Gillis and François Glineur.
\newblock On the geometric interpretation of the nonnegative rank.
\newblock \emph{Linear Algebra and its Applications}, 437\penalty0
  (11):\penalty0 2685 -- 2712, 2012.
\newblock ISSN 0024-3795.

\bibitem[Grais \& Erdogan(2011)Grais and Erdogan]{grais2011single}
Emad~M Grais and Hakan Erdogan.
\newblock Single channel speech music separation using nonnegative matrix
  factorization and spectral masks.
\newblock In \emph{Digital Signal Processing (DSP)}, pp.\  1--6. IEEE, 2011.

\bibitem[Guillamet \& Vitria(2002)Guillamet and Vitria]{guillamet2002non}
David Guillamet and Jordi Vitria.
\newblock Non-negative matrix factorization for face recognition.
\newblock In \emph{Topics in artificial intelligence}, pp.\  336--344.
  Springer, 2002.

\bibitem[Hochreiter \& Schmidhuber(1997)Hochreiter and
  Schmidhuber]{hochreiter1997flat}
Sepp Hochreiter and J{\"u}rgen Schmidhuber.
\newblock Flat minima.
\newblock \emph{Neural Computation}, 9\penalty0 (1):\penalty0 1--42, 1997.

\bibitem[Keskar et~al.(2016)Keskar, Mudigere, Nocedal, Smelyanskiy, and
  Tang]{keskar2016large}
Nitish~Shirish Keskar, Dheevatsa Mudigere, Jorge Nocedal, Mikhail Smelyanskiy,
  and Ping Tak~Peter Tang.
\newblock On large-batch training for deep learning: Generalization gap and
  sharp minima.
\newblock \emph{arXiv preprint arXiv:1609.04836}, 2016.

\bibitem[Kingma \& Ba(2015)Kingma and Ba]{kingma2015adam}
Diederik~P Kingma and Jimmy Ba.
\newblock Adam: A method for stochastic optimization.
\newblock 2015.

\bibitem[Klauck(2003)]{klauck2003rectangle}
Hartmut Klauck.
\newblock Rectangle size bounds and threshold covers in communication
  complexity.
\newblock In \emph{Computational Complexity, 2003. Proceedings. 18th IEEE
  Annual Conference on}, pp.\  118--134. IEEE, 2003.

\bibitem[Krizhevsky \& Hinton(2009)Krizhevsky and
  Hinton]{krizhevsky2009learning}
Alex Krizhevsky and Geoffrey Hinton.
\newblock Learning multiple layers of features from tiny images.
\newblock 2009.

\bibitem[Lee \& Seung(1999)Lee and Seung]{lee1999learning}
Daniel~D Lee and H~Sebastian Seung.
\newblock Learning the parts of objects by non-negative matrix factorization.
\newblock \emph{Nature}, 401\penalty0 (6755):\penalty0 788, 1999.

\bibitem[Li et~al.(2017)Li, Xu, Taylor, and Goldstein]{li2017visualizing}
Hao Li, Zheng Xu, Gavin Taylor, and Tom Goldstein.
\newblock Visualizing the loss landscape of neural nets.
\newblock \emph{arXiv preprint arXiv:1712.09913}, 2017.

\bibitem[Liang et~al.(2017)Liang, Poggio, Rakhlin, and Stokes]{liang2017fisher}
Tengyuan Liang, Tomaso Poggio, Alexander Rakhlin, and James Stokes.
\newblock Fisher-rao metric, geometry, and complexity of neural networks.
\newblock \emph{arXiv preprint arXiv:1711.01530}, 2017.

\bibitem[Morcos et~al.(2018)Morcos, Barrett, Rabinowitz, and
  Botvinick]{morcos2018importance}
Ari~S Morcos, David~GT Barrett, Neil~C Rabinowitz, and Matthew Botvinick.
\newblock On the importance of single directions for generalization.
\newblock \emph{arXiv preprint arXiv:1803.06959}, 2018.

\bibitem[Netzer et~al.(2011)Netzer, Wang, Coates, Bissacco, Wu, and
  Ng]{netzer2011reading}
Yuval Netzer, Tao Wang, Adam Coates, Alessandro Bissacco, Bo~Wu, and Andrew~Y
  Ng.
\newblock Reading digits in natural images with unsupervised feature learning.
\newblock In \emph{NIPS workshop on deep learning and unsupervised feature
  learning}, volume 2011, pp.\ ~5, 2011.

\bibitem[Neyshabur et~al.(2017)Neyshabur, Bhojanapalli, McAllester, and
  Srebro]{neyshabur2017exploring}
Behnam Neyshabur, Srinadh Bhojanapalli, David McAllester, and Nati Srebro.
\newblock Exploring generalization in deep learning.
\newblock In \emph{Advances in Neural Information Processing Systems}, pp.\
  5947--5956, 2017.

\bibitem[Novak et~al.(2018)Novak, Bahri, Abolafia, Pennington, and
  Sohl-Dickstein]{novak2018sensitivity}
Roman Novak, Yasaman Bahri, Daniel~A Abolafia, Jeffrey Pennington, and Jascha
  Sohl-Dickstein.
\newblock Sensitivity and generalization in neural networks: an empirical
  study.
\newblock \emph{ICLR}, 2018.

\bibitem[Russakovsky et~al.(2015)Russakovsky, Deng, Su, Krause, Satheesh, Ma,
  Huang, Karpathy, Khosla, Bernstein, Berg, and Fei-Fei]{ILSVRC15}
Olga Russakovsky, Jia Deng, Hao Su, Jonathan Krause, Sanjeev Satheesh, Sean Ma,
  Zhiheng Huang, Andrej Karpathy, Aditya Khosla, Michael Bernstein,
  Alexander~C. Berg, and Li~Fei-Fei.
\newblock {ImageNet Large Scale Visual Recognition Challenge}.
\newblock \emph{International Journal of Computer Vision (IJCV)}, 115\penalty0
  (3):\penalty0 211--252, 2015.
\newblock \doi{10.1007/s11263-015-0816-y}.

\bibitem[Salamon et~al.(2014)Salamon, Jacoby, and Bello]{salamon2014urban}
J.~Salamon, C.~Jacoby, and J.~P. Bello.
\newblock A dataset and taxonomy for urban sound research.
\newblock In \emph{22nd {ACM} International Conference on Multimedia
  (ACM-MM'14)}, pp.\  1041--1044, Orlando, FL, USA, Nov. 2014.

\bibitem[Simonyan \& Zisserman(2014)Simonyan and Zisserman]{Simonyan2014}
Karen Simonyan and Andrew Zisserman.
\newblock Very deep convolutional networks for large-scale image recognition.
\newblock \emph{arXiv preprint arXiv:1409.1556}, 2014.

\bibitem[Srivastava et~al.(2014)Srivastava, Hinton, Krizhevsky, Sutskever, and
  Salakhutdinov]{srivastava2014dropout}
Nitish Srivastava, Geoffrey Hinton, Alex Krizhevsky, Ilya Sutskever, and Ruslan
  Salakhutdinov.
\newblock Dropout: a simple way to prevent neural networks from overfitting.
\newblock \emph{The Journal of Machine Learning Research}, 15\penalty0
  (1):\penalty0 1929--1958, 2014.

\bibitem[Vapnik(1998)]{vapnik1998statistical}
Vladimir Vapnik.
\newblock \emph{Statistical learning theory. 1998}.
\newblock Wiley, New York, 1998.

\bibitem[Vavasis(2009)]{vavasis2009complexity}
Stephen~A Vavasis.
\newblock On the complexity of nonnegative matrix factorization.
\newblock \emph{SIAM Journal on Optimization}, 20\penalty0 (3):\penalty0
  1364--1377, 2009.

\bibitem[Wang et~al.(2016)Wang, Mohamed, Caruana, Bilmes, Plilipose,
  Richardson, Geras, Urban, and Aslan]{wang2016analysis}
Shengjie Wang, Abdel-rahman Mohamed, Rich Caruana, Jeff Bilmes, Matthai
  Plilipose, Matthew Richardson, Krzysztof Geras, Gregor Urban, and Ozlem
  Aslan.
\newblock Analysis of deep neural networks with extended data jacobian matrix.
\newblock In \emph{International Conference on Machine Learning}, pp.\
  718--726, 2016.

\bibitem[Xiao et~al.(2017)Xiao, Rasul, and Vollgraf]{xiao2017fashion}
Han Xiao, Kashif Rasul, and Roland Vollgraf.
\newblock Fashion-mnist: a novel image dataset for benchmarking machine
  learning algorithms.
\newblock \emph{arXiv preprint arXiv:1708.07747}, 2017.

\bibitem[Xu et~al.(2003)Xu, Liu, and Gong]{xu2003document}
Wei Xu, Xin Liu, and Yihong Gong.
\newblock Document clustering based on non-negative matrix factorization.
\newblock In \emph{Proceedings of the 26th annual international ACM SIGIR
  conference on Research and development in informaion retrieval}, pp.\
  267--273. ACM, 2003.

\bibitem[Zhang et~al.(2016)Zhang, Bengio, Hardt, Recht, and
  Vinyals]{zhang2016understanding}
Chiyuan Zhang, Samy Bengio, Moritz Hardt, Benjamin Recht, and Oriol Vinyals.
\newblock Understanding deep learning requires rethinking generalization.
\newblock \emph{arXiv preprint arXiv:1611.03530}, 2016.

\end{thebibliography}
	\bibliographystyle{iclr2019_conference}
	
	\clearpage
	
	\section{Appendix}
	
	\subsection{Neural network architectures}
	\begin{table}[h]
		\begin{tabular}[t]{lcccc}
			\multicolumn{5}{c}{CIFAR-10} \\
			Type & Out dim. & Kernel & Padding & Stride \\ \hline
			$\text{Conv}_+$ & 64 &  3 &  1 &  1 \\
			$\text{Conv}_+$ & 64 &  3 &  1 &  1 \\
			$\text{Conv}_+$ & 128 &  3 &  1 &  2 \\
			$\text{Conv}_+$ & 128 &  3 &  1 &  1 \\
			$\text{Conv}_+$ & 128 &  3 &  1 &  1 \\
			$\text{Conv}_+$ & 256 &  3 &  1 &  2 \\
			$\text{Conv}_+$ & 256 &  3 &  1 &  1 \\
			$\text{Conv}_+$ & 256 &  3 &  1 &  1 \\
			$\text{Conv}_+$ & 512 &  3 &  1 &  2 \\
			$\text{Conv}_+$ & 512 &  3 &  1 &  1 \\
			Conv & 512 &  3 &  1 &  1 \\
			Linear &  10 &- &- & -\\
		\end{tabular}
		\begin{tabular}[t]{|lcccc}
			\multicolumn{5}{c}{Urban Sounds} \\ 
			Type & Out dim. & Kernel & Padding & Stride \\ \hline
			$\text{Conv}_+$ & 64 &  3 &  1 &  1 \\
			MaxPool & -& 2 & -&1\\
			$\text{Conv}_+$ & 128 &  3 &  1 &  1 \\
			$\text{Conv}_+$ & 128 &  3 &  1 &  1 \\
			MaxPool & -& 2 & -&1\\
			$\text{Conv}_+$ & 256 &  3 &  1 &  1 \\
			$\text{Conv}_+$ & 256 &  3 &  1 &  1 \\
			MaxPool & -& 2 &- & 1\\
			$\text{Conv}_+$ & 512 &  3 &  1 &  1 \\
			$\text{Conv}_+$ & 512 &  3 &  1 &  1 \\
			MaxPool & -& 2 &- & 1\\
			$\text{Linear}_+$ &  4096 &- &- & -\\
			$\text{Linear}_+$ &  4096 & -& -& -\\
			Linear &  10 & -&- &- 
		\end{tabular}\\[2ex]
		
		\begin{tabular}[t]{lcccc|}
			\multicolumn{5}{c}{SVHN} \\
			Type & Out dim. & Kernel & Padding & Stride \\ \hline
			$\text{Conv}_+$ & 64 &  3 &  1 &  1 \\
			$\text{Conv}_+$ & 64 &  3 &  1 &  1 \\
			$\text{Conv}_+$ & 128 &  3 &  1 &  2 \\
			$\text{Conv}_+$ & 128 &  3 &  1 &  1 \\
			$\text{Conv}_+$ & 256 &  3 &  1 &  2 \\
			$\text{Conv}_+$ & 256 &  3 &  1 &  1 \\
			$\text{Conv}_+$ & 512 &  3 &  1 &  2 \\
			Conv & 512 &  3 &  1 &  1 \\
			Linear &  10 &- &- &- \\
		\end{tabular}
		\begin{tabular}[t]{lcccc}
			\multicolumn{5}{c}{Fashion-MNIST} \\
			Type & Out dim. & Kernel & Padding & Stride \\ \hline
			$\text{Linear}_+$ &  128 & -& -& -\\
			$\text{Linear}_+$ &  512 &- &- &- \\
			$\text{Linear}_+$ &  2048 &- &- &- \\
			$\text{Linear}_+$ &  2048 & -&- & -\\
			Linear &  10 & -& -& -\\
		\end{tabular}
		\caption{Neural architecture used for each dataset in Section \ref{sec:experiments}.} \label{tab:arch}
	\end{table}
	
	The exact architectures we used for each dataset are given in Table \ref{tab:arch}. We denote a linear or convolutional layer followed by a ReLU as $\text{Linear}_+$ and $\text{Conv}_+$, respectively.

	\subsection{Ablating NMF and PCA directions}
	It is interesting to study the impact of \emph{ablating} the activation in the directions found by NMF and PCA by forward propagating the residual, i.e., \begin{align}
	\mA_{i+1} = \max \left( (\mA_k - \tilde{\mA}_k)W_{i+1}, 0 \right) \label{eq:residual}
	\end{align}
	
	This is interesting because in the case of PCA, for instance, the top $k$ directions are those that capture most of the variance in the activation matrix, and presumably the $k$ directions found by NMF are of similar importance. This is not true for the random ablations, where the ablated directions are of no special importance.
	
\begin{figure}
		\centering
		\bgroup
		\setlength{\tabcolsep}{0.2pt}
		\begin{tabular}{cccc}
			& CIFAR-10 & Urban Sounds & Fashion-MNIST \\
			& \multicolumn{3}{c}{Single-class NMF}  \\
			\rotatebox[origin=c]{90}{\centering Accuracy} \hspace{0.3pt} &
			\raisebox{-.5\height}{\includegraphics[width=.32\textwidth]{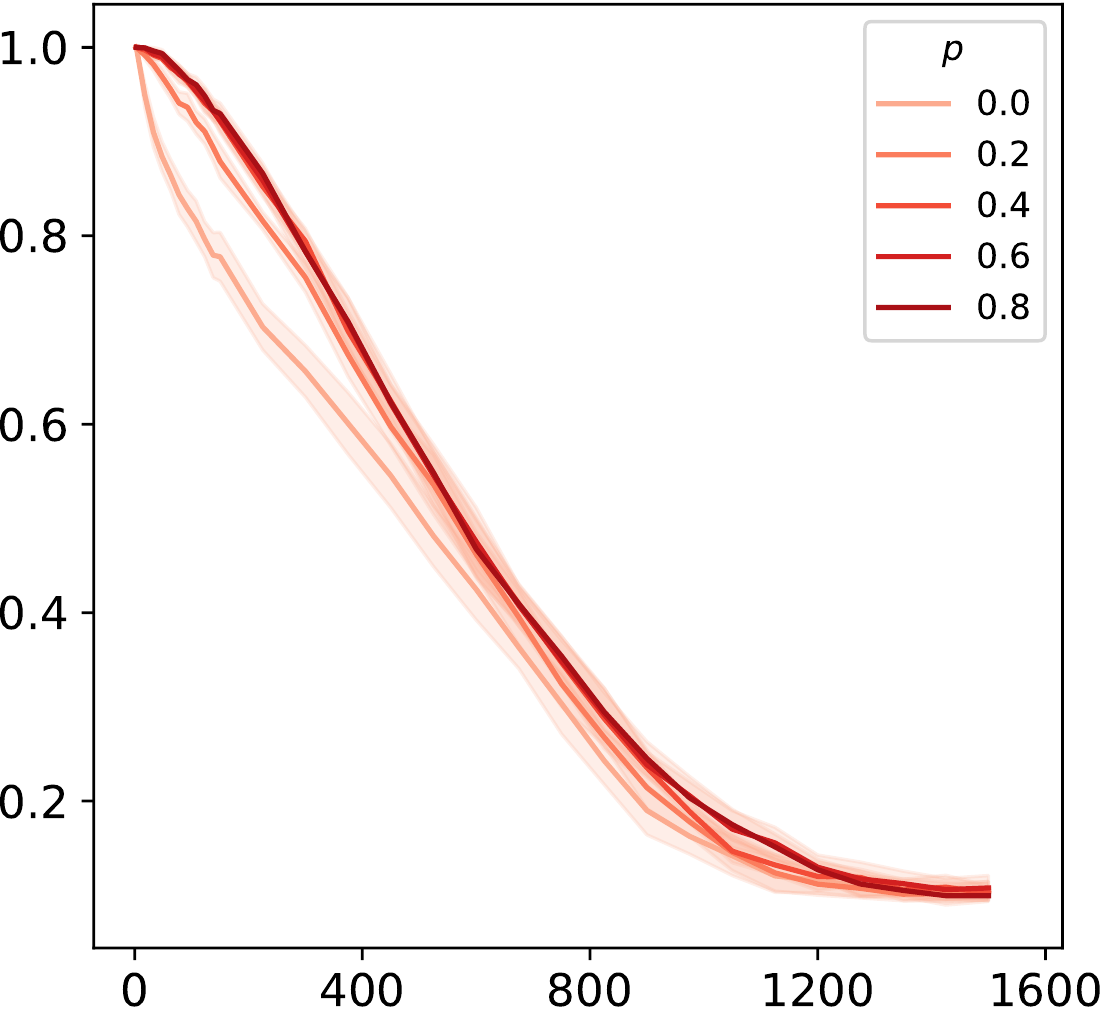}}&
			\raisebox{-.5\height}{\includegraphics[width=.32\textwidth]{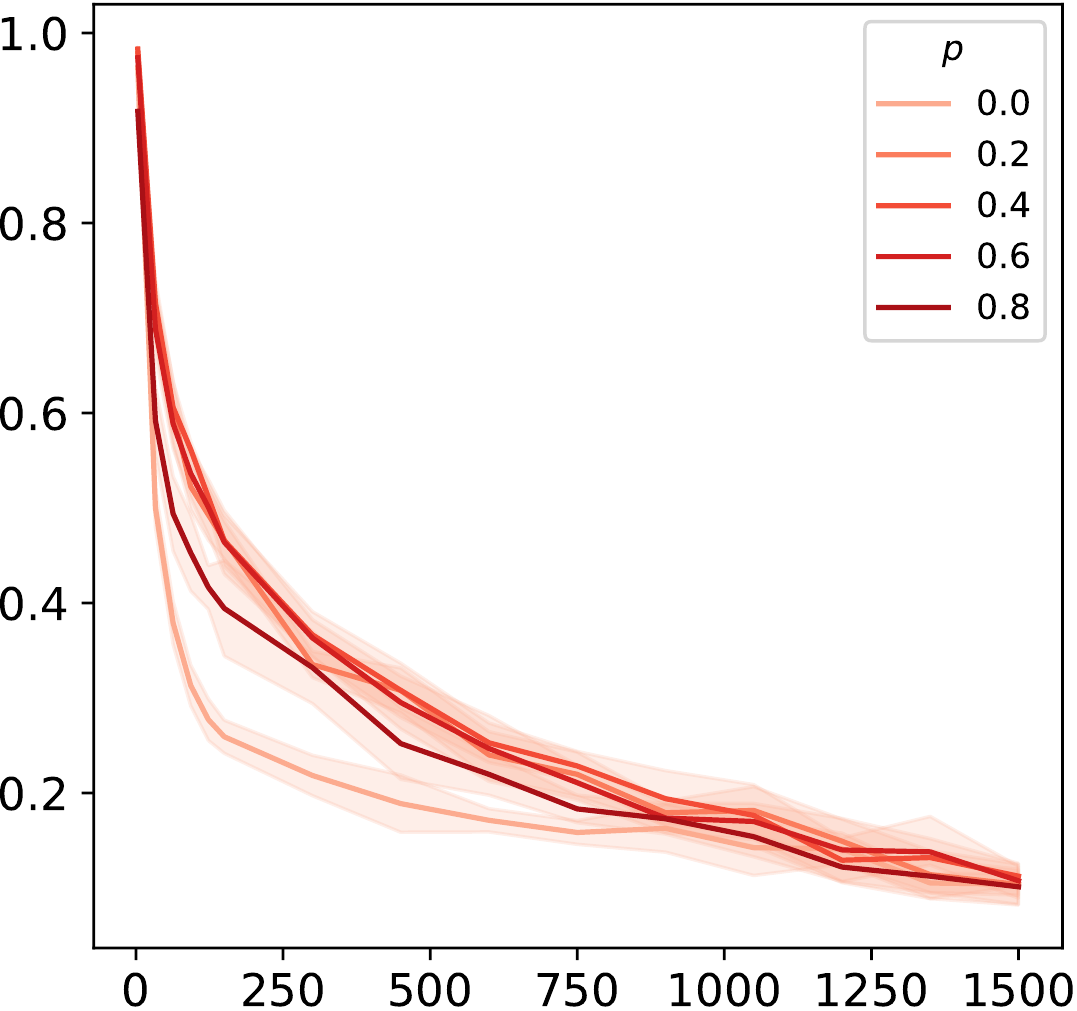}}&
			\raisebox{-.5\height}{\includegraphics[width=.32\textwidth]{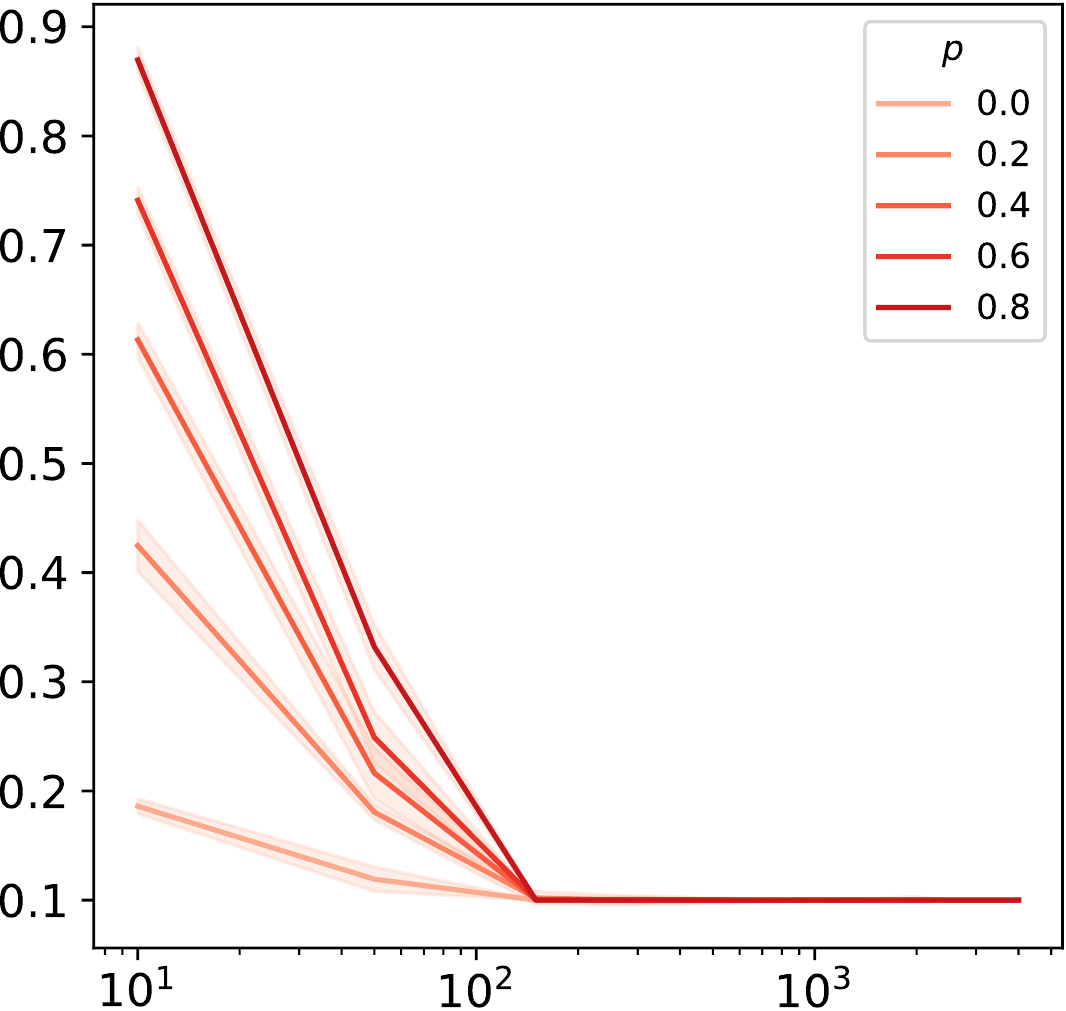}}  \vspace{2pt}\\
			& \multicolumn{3}{c}{Number of dimensions $k$} 
			\\
			& \multicolumn{3}{c}{Single-class PCA}  \\
			\rotatebox[origin=c]{90}{\centering Accuracy} \hspace{1.0pt} &
			\raisebox{-.5\height}{\includegraphics[width=.315\textwidth]{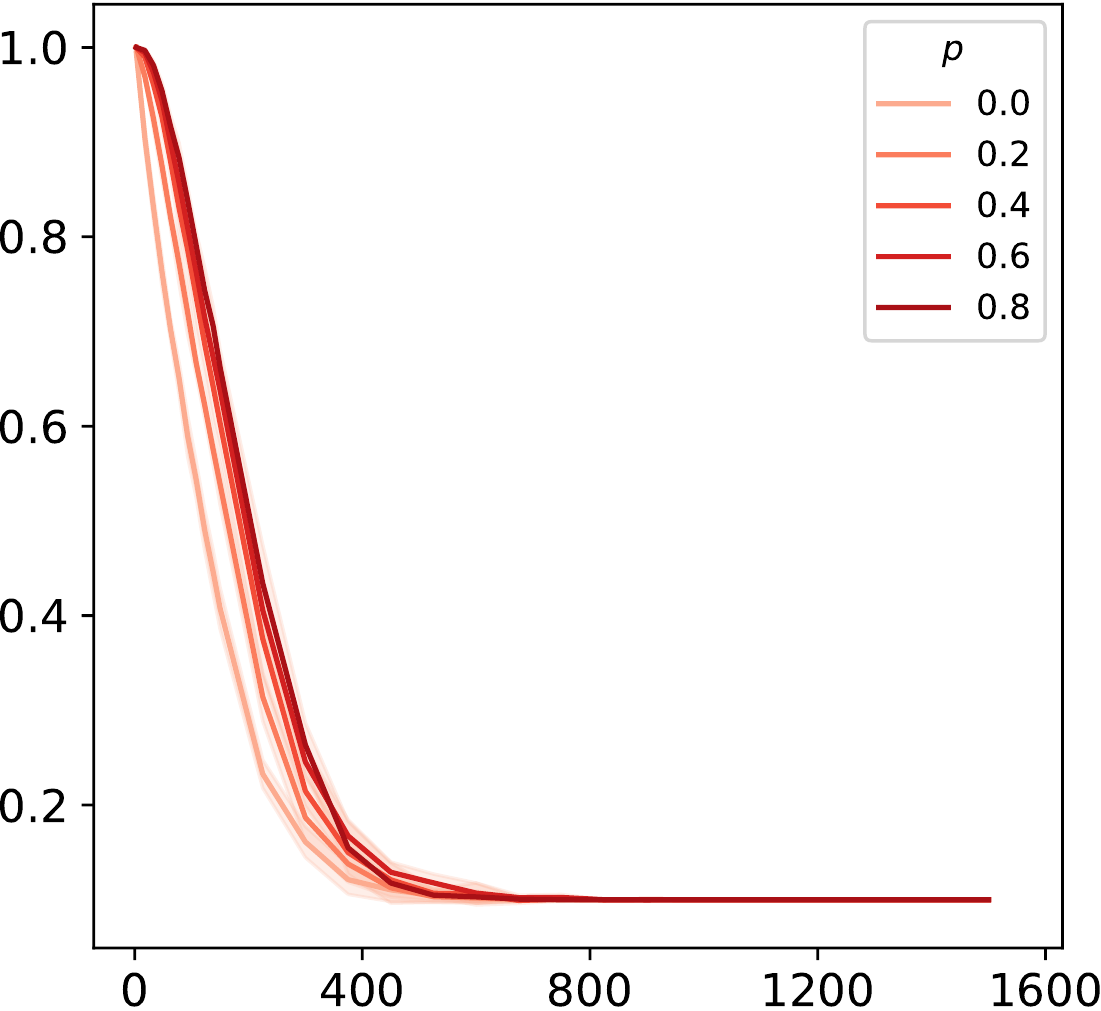}}&
			\raisebox{-.5\height}{\includegraphics[width=.315\textwidth]{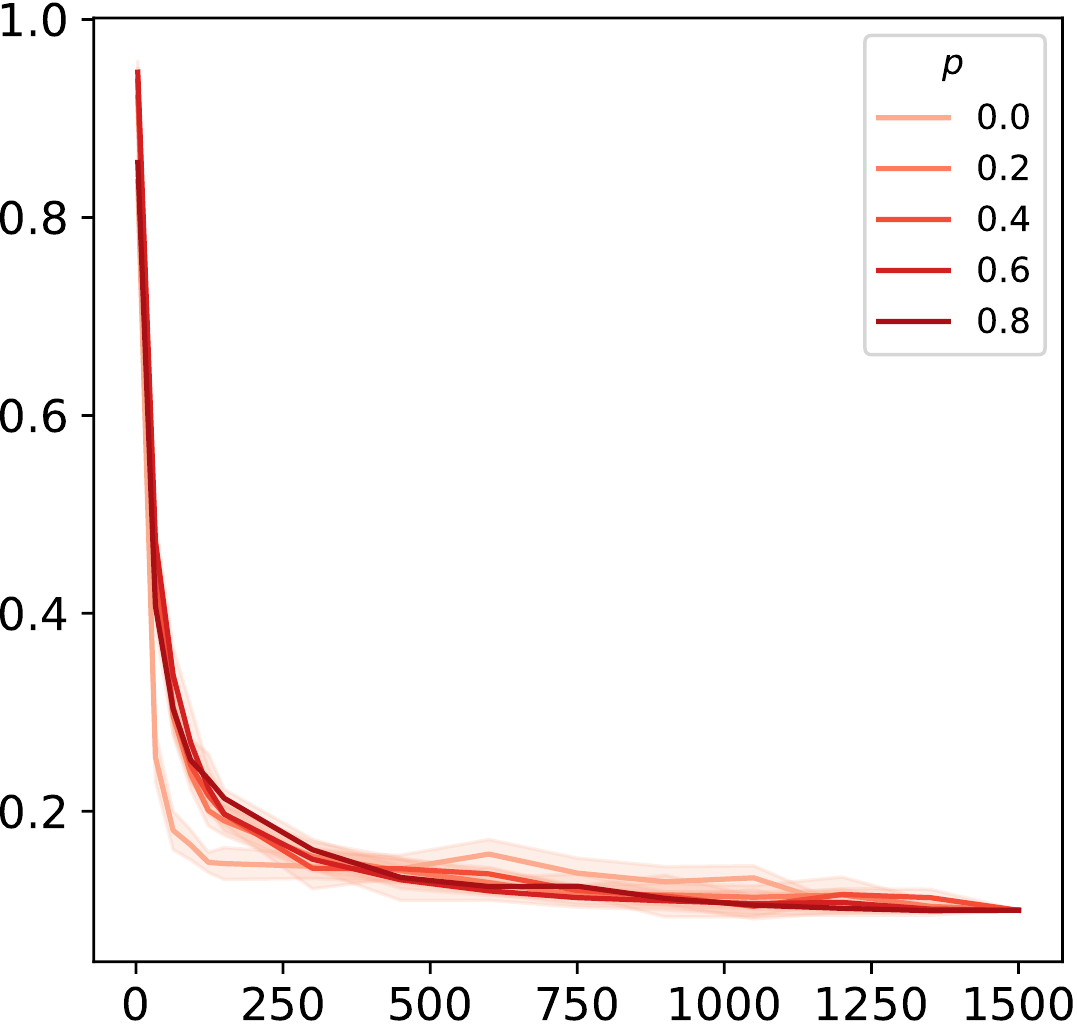}}&
			\raisebox{-.5\height}{\includegraphics[width=.315\textwidth]{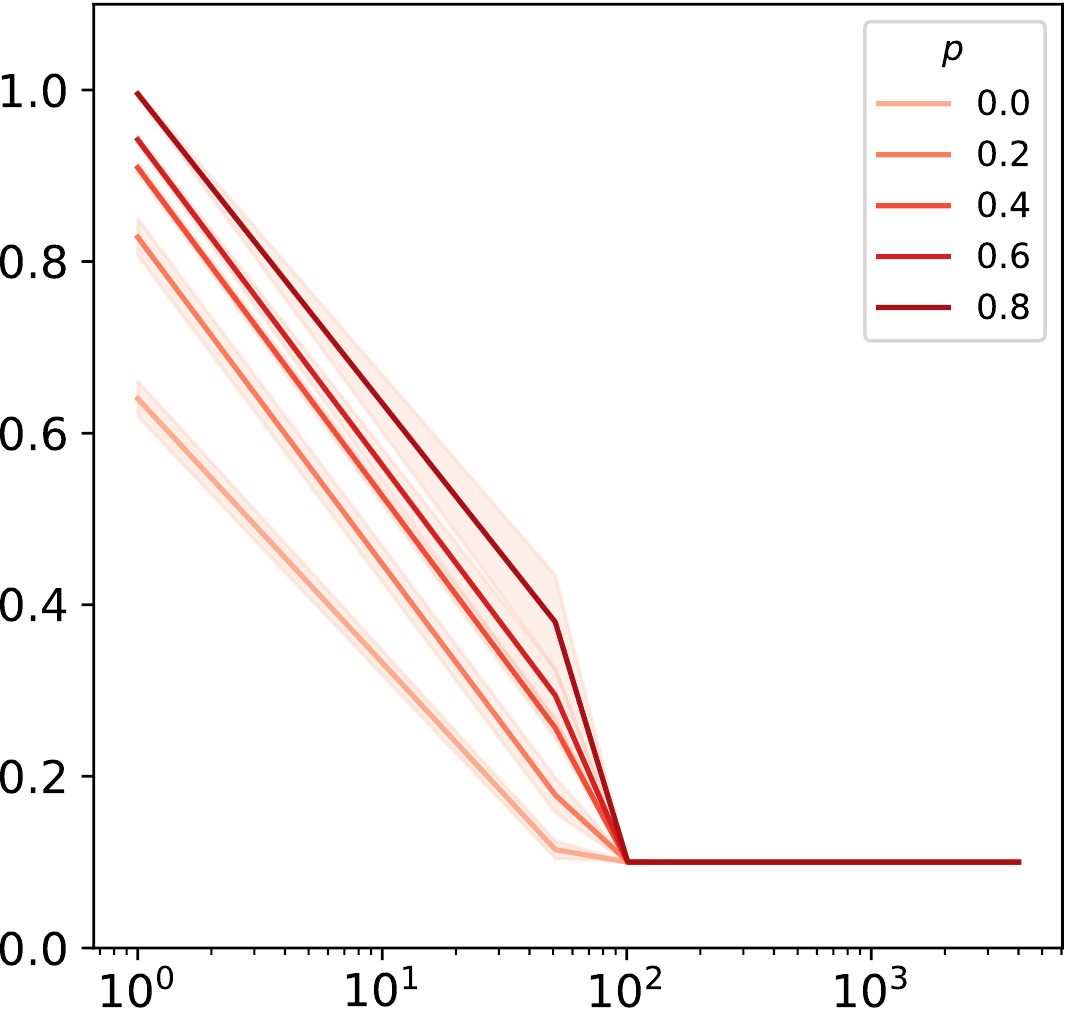}}  \vspace{2pt} \\
			& \multicolumn{3}{c}{Number of dimensions $k$}
		\end{tabular}
		\egroup
		\caption{
			\textbf{NMF and PCA directions are more important} for networks with induced-memorization. Each column shows results for a specific dataset and network architecture. Compared to random ablation as in Figures~\ref{fig:CIFAR} and \ref{fig:mnist}, where networks with induced-memorization are most robust, in all cases we see that ablation in NMF and PCA directions hurts their performance more, compared to memorizing networks.
		}
		\label{fig:residuals}
	\end{figure}

	In Figure \ref{fig:residuals}  we see that networks with no induced memorization that are \emph{most vulnerable} to ablation of NMF and PCA direction. In other words, while non-memorizing networks are more robust to \emph{random} ablations, they are not robust to ablations of specific important directions. This is in contrast to the interpretation of \citet{morcos2018importance} that non-memorizing networks are more robust to ablations of single directions.

	\subsection{NMF on VGG-19} \label{sec:vgg}
	
	The VGG-19 model~\citep{Simonyan2014}, trained on ImageNet~\citep{ILSVRC15}, is known for its good generalization ability, as evident by its widespread use as a general feature extractor. We use a pre-trained model here as an example of a well-generalizing network and analyze it with our method.
	
	We apply NMF compression to the three deepest convolutional layer, on activations of both single-class batches and multi-class batches. We select 50 random classes from ImageNet and gather batches of 50 training samples from each class.
	
\begin{figure}
		\centering
		\bgroup
		\setlength{\tabcolsep}{0.2pt}
		\begin{tabular}{cccc}
			& (a) Single vs. multi class & (b) Residuals & (c) Per-class test accuracy \\
			\rotatebox[origin=c]{90}{\centering Accuracy} \hspace{0.3pt} &
			\raisebox{-.5\height}{\includegraphics[width=.32\textwidth]{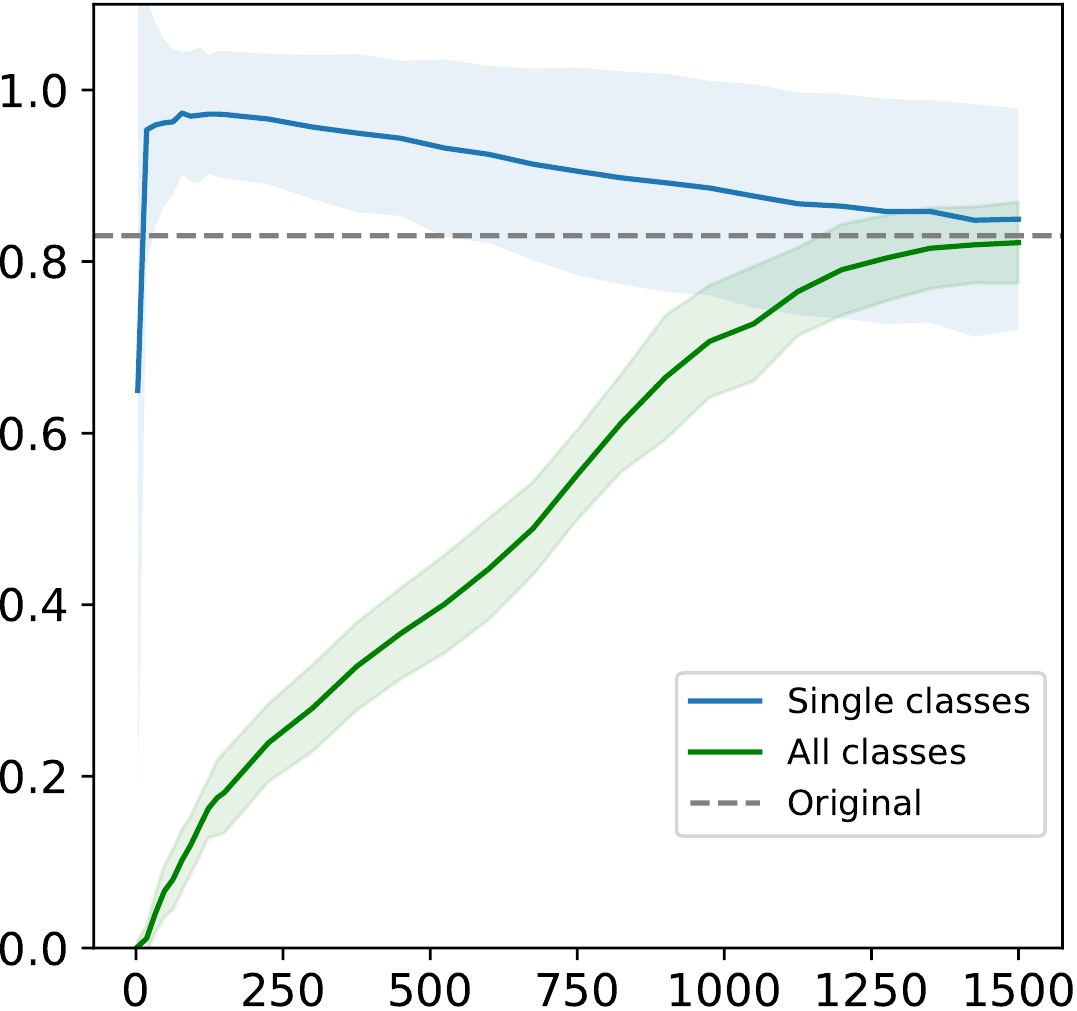}}&
			\raisebox{-.5\height}{\includegraphics[width=.32\textwidth]{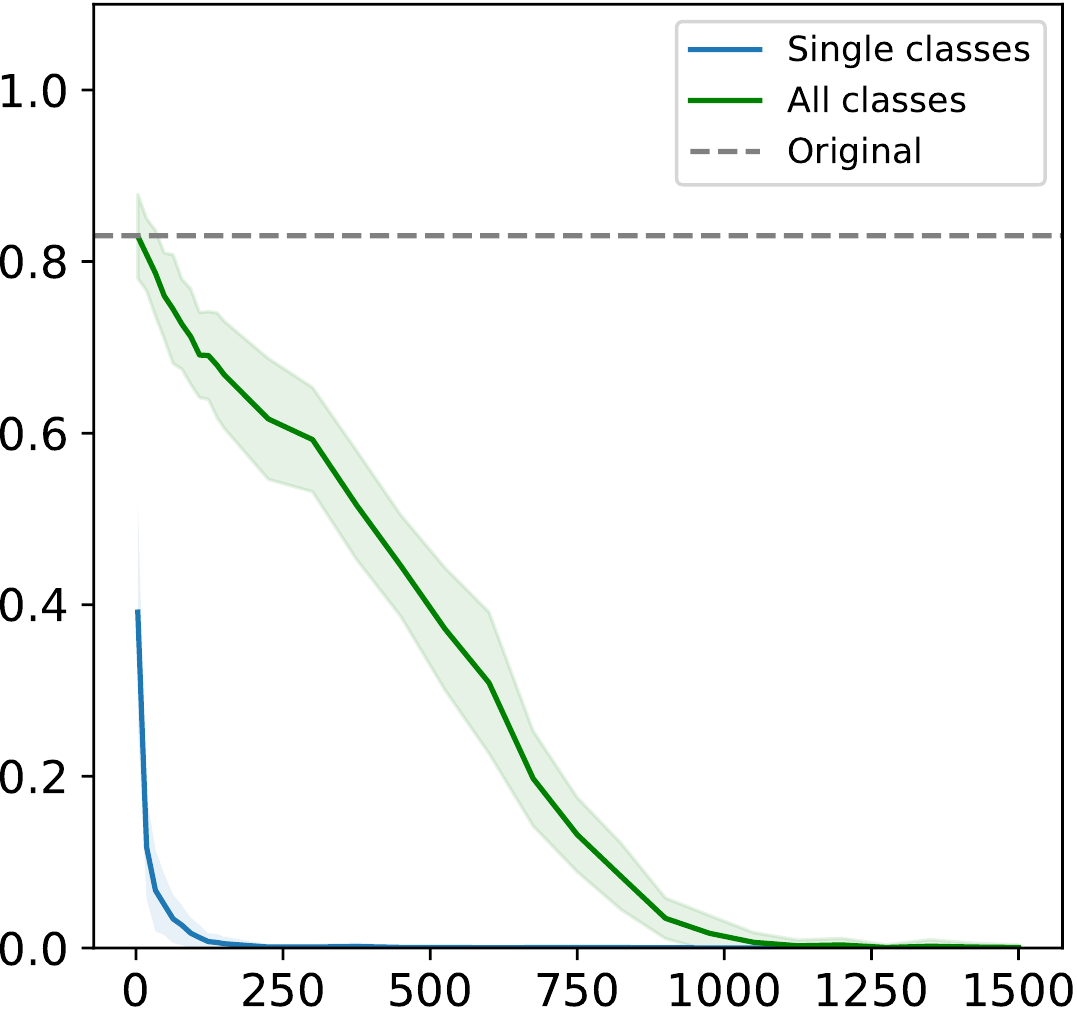}}&
			\raisebox{-.5\height}{\includegraphics[width=.32\textwidth]{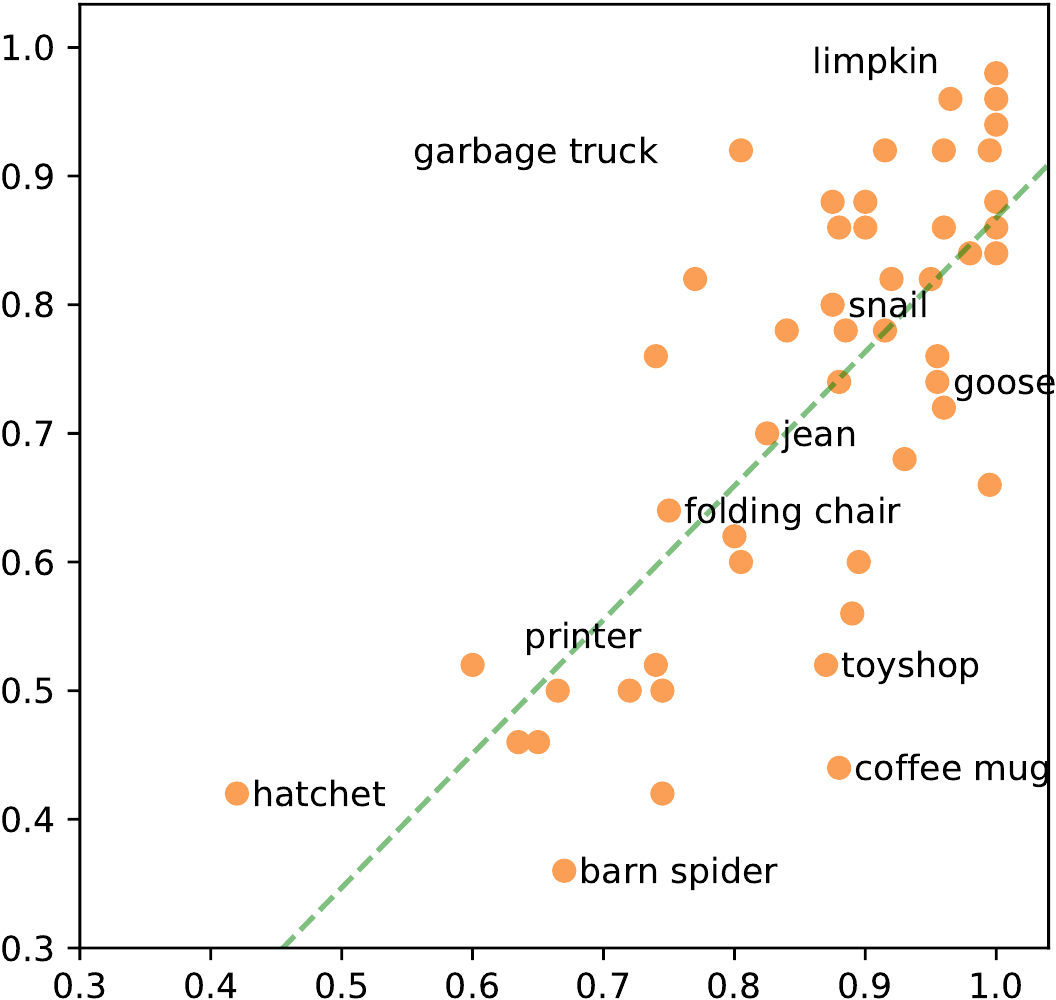}}  \vspace{2pt}\\
			& \multicolumn{2}{c}{Non-negative rank $k$} & NMF AuC
		\end{tabular}
		\egroup
		\caption{
			\textbf{NMF compression on VGG-19}.
			(a) Deep VGG layers are highly linear with respect to single-class batches, as indicated by high accuracy for small dimensions of $k$. Compression can have a denoising effect and improve upon the baseline accuracy of the batch (dashed line). (b)
			Removing NMF directions causes a dramatic drop in accuracy, more so on single-class batches.
			(c)  Per-class test set accuracy  is significantly correlated with the area under the $k$ vs. accuracy curve (NMF AuC).
		}
		\label{fig:vgg}
	\end{figure}

	In Figure~\ref{fig:vgg}, shown in blue, NMF applied to single-class batches, has a denoising effect and improves over the baseline accuracy of the batch, shown as a dashed line. As the constraint on $k$ is relaxed, that accuracy drops back to its baseline level.  
	
	We contrast this behavior with the one shown in green when using multi-class batches. Here, only when $k$ is large do we regain baseline accuracy, and sensitivity to ablation is similarly diminished. This is due to the critical role non-linearity plays in separating the different classes.
	Ablating the NMF directions dramatically reduces classification accuracy.
	Finally, in Figure~\ref{fig:vgg} (c) we show there is a significant per-class correlation (Pearson $r=0.78$ ) between NMF AuC and accuracy on test accuracy on batches from the ImageNet test set.

\end{document}